%% file: example.tex
\documentclass{article}

\usepackage[final]{corl_2026}
\usepackage{graphicx}
\usepackage{booktabs}
\usepackage{hyperref}
\usepackage{algorithm}
\usepackage{algpseudocode}
\usepackage{amsmath,amssymb}
\usepackage{wrapfig}
\usepackage{amsfonts}
\usepackage{capt-of}
\usepackage{orcidlink}
\usepackage{pifont}
\newcommand{\Letter}{\ding{41}}
\usepackage{float} 
\usepackage{comment}
\usepackage{multirow}   
\usepackage{xcolor}     
\usepackage[table]{xcolor}

\title{OptiGeo: Efficient Monocular Geometry for Embodied Perception in Optically Challenging Scenes}

\author{
\normalfont
Muxin Liu$^{1,2,*}$ \quad
Tianbo Liu$^{1,*}$ \quad
Jing Xia$^{1,*}$ \quad
Xiaoyang Lyu$^{1}$ \quad
Xiaoshan Wu$^{1}$ \quad
Bo Wang$^{1}$ \quad \\[1pt]
Peng Dai$^{1}$ \quad
Zhongrui Wang$^{3}$ \quad
Shaoshuai Shi$^{2}$\textsuperscript{\Letter} \quad
Xiaojuan Qi$^{1}$\textsuperscript{\Letter} \\[4pt]
$^{1}$The University of Hong Kong \qquad
$^{2}$Voyager Research, DiDi Chuxing \\
$^{3}$Southern University of Science and Technology \\[3pt]
$^{*}$Equal contribution
\qquad
\Letter~Corresponding author \\[3pt]
\texttt{mxliu@connect.hku.hk, timber@connect.hku.hk, } 
\texttt{jing.xia-666@connect.hku.hk} \\
\texttt{shaoshuaics@gmail.com, xjqi@eee.hku.hk}
}

\begin{document}
\maketitle

\vspace{-20pt}
\begin{center}
    \centering
    \includegraphics[width=0.99\textwidth]{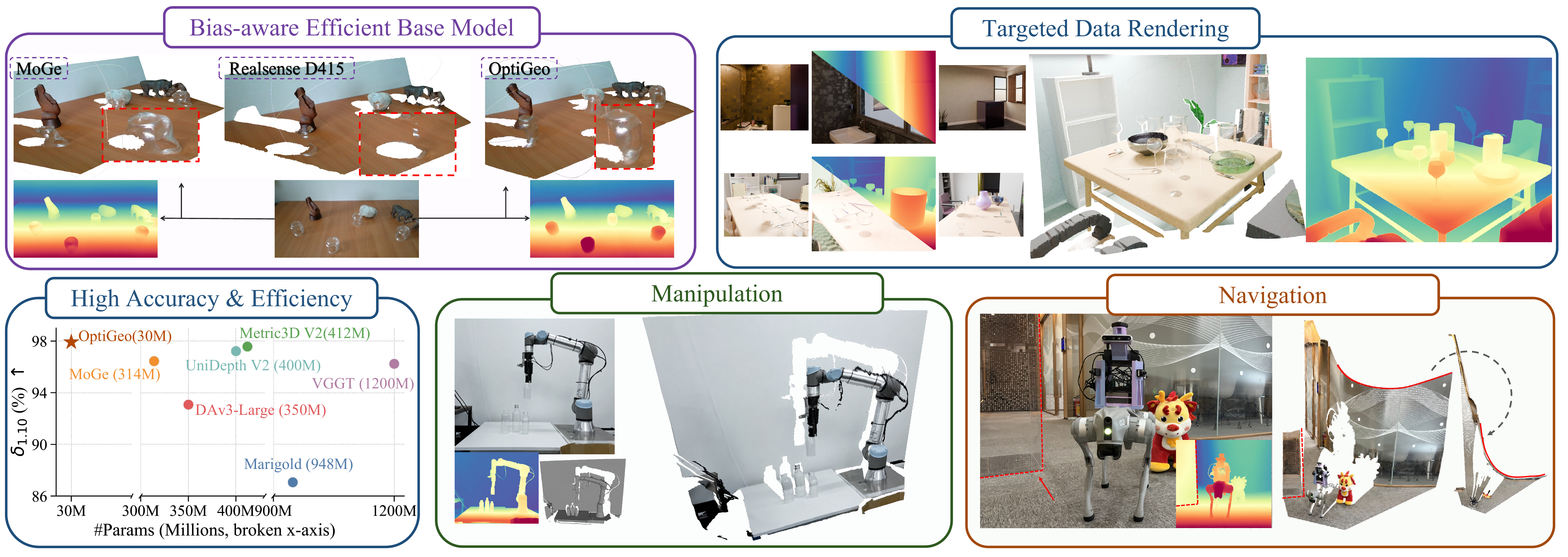}
    \captionof{figure}{Overview of OptiGeo: a compact geometry model with robust transparent-object reconstruction, trained with rendered data and validated on downstream tasks.}
    \label{fig:overview}

\end{center}%
%===============================================================================

\input{sec/0_abstract}
\input{sec/1_introduction}
\input{sec/2_related_work}
\input{sec/3_empirical_study}
\input{sec/4_methods}
\input{sec/5_experiments}

\input{sec/6_conclusion}

%===============================================================================

\bibliography{example}  % .bib

\input{sec/X_suppl}

\end{document}

%% file: sec/0_abstract.tex
\begin{abstract} 
Monocular depth estimation has achieved strong open-domain generalization, yet reliable robotic deployment remains difficult in transparent, reflective, and specular environments, where depth sensors often produce missing or biased depth. 
Existing methods often handle such optical failures with scene-specific preprocessing, auxiliary modules, or post-hoc fine-tuning. While effective in constrained settings, these designs increase architectural redundancy and can over-specialize general geometry models to narrow optical scenarios. 
We revisit this problem as a localized failure mode within base-model training and identify sensor-induced supervision bias as a key bottleneck: models inherit sensor failure patterns from biased real-depth supervision in optically challenging regions.
We then introduce \textbf{OptiGeo}, a bias-aware training framework that rehabilitates biased real supervision using a clean-geometry teacher and residual-trimmed alignment. We redefine transparency-targeted rendering as a compact source of clean optical geometry, rather than a large domain-specific fine-tuning set. With only a small targeted rendering set, OptiGeo learns the geometric structure of transparent objects and regions, correcting local geometry distortions that real sensors cannot reliably supervise.
Despite only 30M parameters, OptiGeo outperforms substantially larger 300M-scale monocular models and billion-scale multi-view baselines on transparent-scene benchmarks, while remaining competitive on general zero-shot depth and boundary sharpness. Real-world navigation cases further validate its practicality as an efficient perception module in optically challenging scenes.
\noindent\textbf{Project page:} \url{https://mx-liu6.github.io/OptiGeo-web/}
\vspace{-6pt}

\end{abstract}

\keywords{Monocular Geometry, Optical Robustness, Embodied Perception}

%% file: sec/1_introduction.tex
\section{Introduction}

\begin{wrapfigure}{r}{0.42\textwidth}
\centering
\vspace{-16pt}
\includegraphics[width=0.99\linewidth]{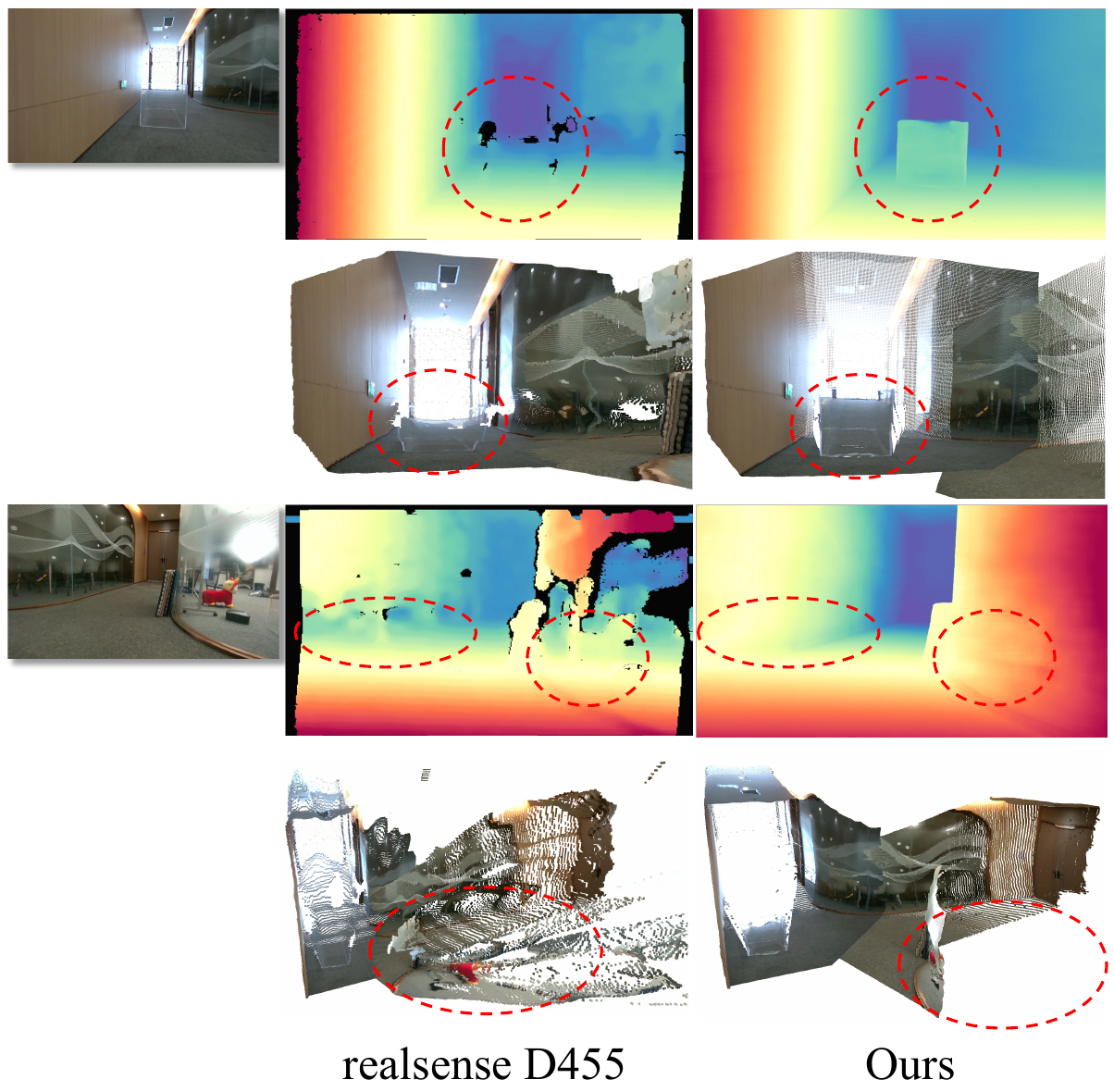}
\vspace{-16pt}
\caption{Transparent-object scenes where real sensor depth fails, while OptiGeo recovers coherent object and surrounding glass geometry.}
\vspace{-14pt}
\label{fig:real_transparent_object}
\end{wrapfigure}

Reliable geometry perception is essential for embodied agents operating in real worlds. 
Navigation systems must reason about glass doors, mirrors, and reflective obstacles, while manipulation systems require accurate object geometry for grasping transparent containers, glassware, and specular objects. 
However, these regions remain challenging for RGB-D and Time-of-Flight sensors: refraction, reflection, and transmission often produce missing, distorted, or background-leaking depth measurements, as shown by the erroneous depth patterns in Fig.~\ref{fig:real_transparent_object}.
When directly used, the transparent object is almost absent. Such failures can change the perceived scene topology by turning physical obstacles into missing or free-space regions.

Recent monocular depth and geometry models have achieved strong open-domain generalization~\cite{liu2026monocular} through stronger backbones~\cite{oquab2024dinov2,simeoni2025dinov3,rombach2021highresolution,lyu2021hr,gao2025more3dvisualgeometry}, large-scale training~\cite{yang2024depth, yang2024depth2, piccinelli2024unidepth, piccinelli2025unidepthv2, hu2024metric3d,liu2026foundationgeo}, and richer geometric representations~\cite{wang2025moge, wang2025moge2, bochkovskiydepth}. 
At the same time, transparent-object perception has often been treated as a separate downstream task, addressed by depth restoration from corrupted RGB-D inputs~\cite{sajjan2020clear,fang2022transcg,dai2022domain}, transparent-object-specific pseudo-label generation~\cite{costanzino2023learning, wang2026seeclearreliabletransparentobject,wang2024digging} or generative/diffusion-based refinement~\cite{xu2025diffusion}.
While effective in constrained settings, these designs may increase system complexity and over-specialize the model to narrow optical domains. 
In contrast, we argue that transparent and reflective regions are better understood as localized, object-level geometry failures within general monocular geometry estimation, rather than as a standalone depth-estimation problem.

Our empirical study in Sec.~\ref{empirical_study} shows that optical failures largely stem from biased real-depth supervision: transparent and reflective regions can contain erroneous yet seemingly valid labels, causing base models to inherit sensor failure patterns (as shown in Fig.~\ref{fig:empirical}).
Since these errors are typically localized and do not dominate scene-level accuracy, the key is to correct biased supervision while preserving the base model's generalization.

We then introduce \textbf{OptiGeo}, a bias-aware training schema for efficient monocular geometry estimation in optically challenging robotic environments. 
We first curate a broad 9M-frame corpus from 21 datasets, including 2.6M synthetic frames from diverse domains. These existing synthetic data provide broad scene-level geometry diversity, but they are not designed to densely cover transparent objects and reflective surfaces.
Therefore, we further construct a compact transparency-targeted rendering set built on Infinigen~\cite{raistrick2023infinite}.
Rather than simply increasing synthetic data volume, this targeted set provides clean object-centric and region-level geometry for optically challenging regions that real sensors cannot reliably annotate.

Based on this curated data, we first train a clean-geometry teacher only on synthetic scenes, so that it provides an optical-aware prior without inheriting real sensor failure patterns.
For real images, residual-trimmed alignment calibrates the teacher prediction to reliable sensor pixels while suppressing high-residual biased regions.
In this way, real data still provides realistic appearance, scene layout, and metric anchoring, while locally corrupted optical-region labels are replaced by corrected supervision.
Finally, the corrected real labels and targeted rendered samples jointly supervise a compact 30M student model within a unified base model architecture.

Experiments show that OptiGeo achieves a strong balance between optical robustness, generalization, and efficiency. 
Despite only 30M parameters, OptiGeo outperforms substantially larger 300M-scale monocular models and billion-scale multi-view baselines on transparent-scene benchmarks. 
It remains competitive on general zero-shot depth evaluation and achieves strong boundary sharpness, indicating accurate fine-structure recovery around transparent and reflective regions. 
Real-world navigation-oriented experiments further validate OptiGeo as an efficient perception module for embodied agents operating in optically challenging scenes.

%% file: sec/2_related_work.tex
\vspace{-4pt}
\section{Related Work} 
\vspace{-4pt}

\vspace{-4pt}
\vspace{0.02in}\noindent \textbf{Open-domain monocular depth and geometry estimation.}
Recent monocular depth estimation has evolved from dataset-specific prediction to open-domain depth and geometry modeling. 
DPT~\cite{ranftl2021vision} introduces transformer backbones for stronger global reasoning, while Depth Anything~\cite{yang2024depth} and Depth Anything V2~\cite{yang2024depth2} leverage large-scale diverse supervision to achieve strong zero-shot relative depth generalization. 
Marigold~\cite{ke2025marigold} shows that diffusion models encode useful geometric priors. 
Metric methods further resolve scale ambiguity through camera-aware modeling or geometric constraints, including ZoeDepth~\cite{bhat2023zoedepth}, Metric3D~\cite{yin2023metric3d}, Metric3D v2~\cite{hu2024metric3d}, UniDepth~\cite{piccinelli2024unidepth}, UniDepth V2~\cite{piccinelli2025unidepthv2} and DepthPro~\cite{bochkovskiydepth}. 
Recent geometry models further improve metric estimation: MoGe-2~\cite{wang2025moge2} adopts relative-to-metric learning for fine-grained geometry recovery, while FoundationGeo~\cite{liu2026foundationgeo} reveals camera-intrinsic distribution mismatch as a key bottleneck for zero-shot metric generalization, and introduces pixel-wise scale and ray-direction fields together with camera-diverse training to improve cross-camera robustness.

\vspace{-4pt}
\vspace{0.02in}\noindent \textbf{Transparent and reflective object perception.}
Transparent and reflective objects remain challenging because optical effects often make RGB-D, ToF, and stereo depth incomplete or biased~\cite{lu2025uniugp,pei2026advancing,jiang2026wpt,tan2025xtrack,lin2024sam}.
ClearGrasp~\cite{sajjan2020clear} predicts transparent masks, surface normals, and occlusion boundaries to refine corrupted sensor depth for robotic manipulation.
TransCG~\cite{fang2022transcg} extends this RGB-D restoration paradigm with real-world transparent-object data for depth completion and grasping.
DREDS/STD~\cite{dai2022domain} simulates sensor failures on specular and transparent objects to improve restoration under optical artifacts.
Depth4ToM~\cite{costanzino2023learning} generates pseudo labels through image inpainting and depth prediction.
MODEST~\cite{liu2025monocular} and D4RD~\cite{wang2024digging} improve monocular transparent-depth estimation with specialized training.
DKT~\cite{xu2025diffusion} leverages diffusion priors for transparent-scene geometry.
While effective, these methods often rely on task-specific preprocessing, auxiliary cues, generative refinement, or post-hoc transparent-scene adaptation.
\vspace{-6pt}

% \vspace{0.05in}\noindent \textbf{Efficient depth perception for robotics.}
% Robotic navigation and manipulation require depth perception that is not only accurate, but also low-latency and robust under domain shift. 
% Efficient monocular depth methods such as FastDepth~\cite{wofk2019fastdepth}, GuideDepth~\cite{rudolph2022lightweight}, and Lite-Mono~\cite{zhang2023lite} reduce computation through lightweight encoders, compact decoders, or efficient feature fusion, making dense prediction more suitable for embedded deployment. 
% However, these models are typically designed for generic indoor or outdoor depth estimation and do not explicitly address optically challenging regions where sensor supervision is biased. 
% For closed-loop robotics, this creates a tension between geometric robustness and real-time deployability. 

%% file: sec/3_empirical_study.tex
\vspace{-4pt}
\section{Empirical Study}
\vspace{-4pt}
\label{empirical_study}

We revisit optical-region failures as a base-model training issue rather than a post-hoc transparent-depth task. While modern monocular geometry models generalize well at the scene level, transparent and reflective regions remain unreliable when supervision inherits real-sensor bias. This motivates learning optical robustness within general depth training by correcting biased supervision, instead of relying on scene-specific preprocessing or large-scale adaptation.

We empirically study three questions: 
(1) how sensor-induced optical bias enters base-model training; 
(2) whether these errors are localized or dominate whole-scene depth accuracy; and 
(3) how transparency-targeted synthetic data should be designed and used. 
The resulting findings motivate our data-driven strategy.

\vspace{-8pt}
\subsection{How Optical Bias Enters Base-Model Training}
\vspace{-6pt}

Depth base models are trained from mixtures of synthetic and real data. 
\begin{wrapfigure}{r}{0.44\textwidth}
    \centering
    \vspace{-14pt}
    \includegraphics[
        width=0.98\linewidth,
        height=0.32\textheight,
        keepaspectratio
    ]{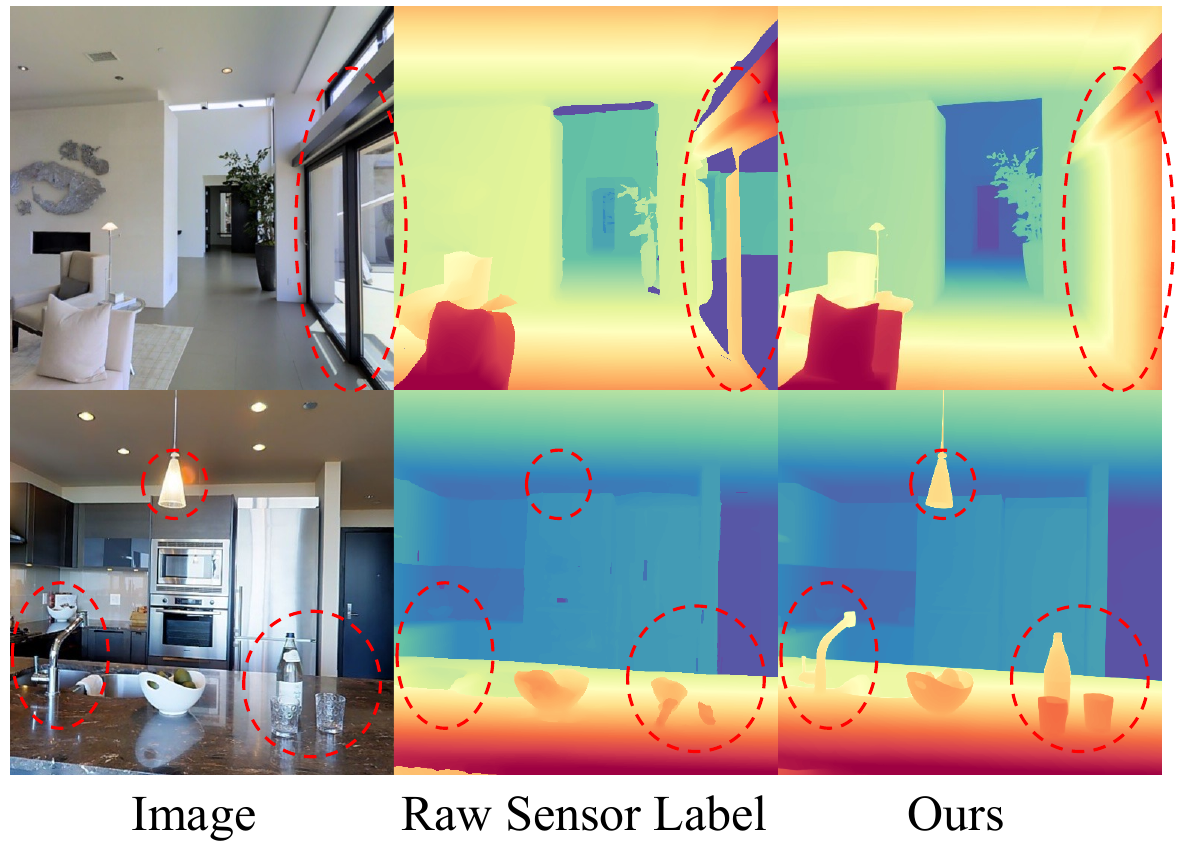}
    \vspace{-8pt}
    \caption{Sensor-induced supervision bias.}
    \label{fig:empirical}
    \vspace{-8pt}
\end{wrapfigure}
Synthetic data provides clean geometric labels, while real data often comes from RGB-D, ToF, stereo, LiDAR, or reconstruction pipelines whose measurements can be unreliable around transparent, reflective, and specular surfaces. 
Thus, optical bias mainly enters through the supervision signal rather than the RGB input. Fig.~\ref{fig:empirical} illustrates this issue. 
Raw sensor labels can assign background depth to glass surfaces, erase transparent objects, or distort reflective and specular regions while still appearing locally valid. Such errors often survive standard preprocessing because they are not simply missing pixels or obvious outliers. 
When used as training labels, they encourage base models to learn sensor failure patterns in the very regions where robust geometry is needed. 
This suggests that transparent and reflective failures are not only a capacity issue, but also a supervision-bias issue that should be corrected before base-model training.

\vspace{-4pt}
\subsection{Are Optical Challenges a Separate Depth Task?}
\vspace{-4pt}

We next examine whether transparent and reflective regions should be treated as a separate depth-estimation task. 
In embodied scenarios, optical failures are often spatially localized: navigation errors typically occur around glass doors, windows, mirrors, or reflective obstacles, while manipulation errors concentrate on transparent containers, specular objects, and their boundaries. 
These failures are important for decision making, but they do not necessarily corrupt the global geometry of the entire scene.

\begin{wraptable}{r}{0.46\textwidth}
    \centering
    \vspace{-16pt}
    \caption{ClearGrasp Real full-image results.}
    \label{tab:cleargrasp_small}
    \scriptsize
    \resizebox{\linewidth}{!}{
    \begin{tabular}{lccc}
        \toprule
        Method & Params & AbsRel$\downarrow$ & $\delta_{1.05}\uparrow$ \\
        \midrule
        DAv3-Large & 350M & 2.94 & 85.6 \\
        VGGT & 1.2B & 1.98 & 91.9 \\
        OptiGeo-Teacher & 940M & \textbf{1.46} & \textbf{94.6} \\
        \bottomrule
    \end{tabular}
    }
    \vspace{-12pt}
\end{wraptable}
As shown in Tab~\ref{tab:cleargrasp_small}, even without a transparency-specific post-hoc pipeline, our teacher model achieves strong full-image performance on standard depth benchmarks. 
This suggests that base-model training is sufficient to recover reliable whole-scene geometry, while the remaining difficulty lies in local optical structures. Thus, transparent and reflective regions should be treated as localized failure cases within general monocular geometry estimation, not as a separate downstream task.

\vspace{-6pt}
\subsection{What Role Should Transparency-targeted Data Play?}
\vspace{-4pt}

The analysis above also changes how transparency-targeted data should be used. Rather than treating it as a large domain-specific fine-tuning set, we use it as a compact source of clean optical geometry that real sensors cannot reliably annotate.
\begin{wrapfigure}{r}{0.48\textwidth}
    \centering
    \vspace{-14pt}
    \includegraphics[
        width=0.98\linewidth,
        height=0.32\textheight,
        keepaspectratio
    ]{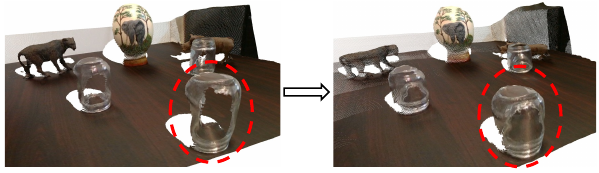}
    \vspace{-8pt}
    \caption{Transparency-targeted rendering helps correct local geometry distortions.}
    \label{fig:target_data}
    \vspace{-8pt}
\end{wrapfigure}
Rendered data provides physically consistent supervision for transparent surfaces, reflective objects, and optical boundaries, exposing the base model to local structures that are often corrupted or missing in real sensor labels.
As illustrated in Fig.~\ref{fig:target_data}, the purpose of targeted rendering is not to expand a separate transparent-scene domain, but to correct local geometry distortions around transparent objects and reflective regions. 

%% file: sec/4_methods.tex
\vspace{-8pt}
\section{OptiGeo}
\vspace{-8pt}

\begin{figure}[t]
    \centering
    \includegraphics[width=\textwidth]{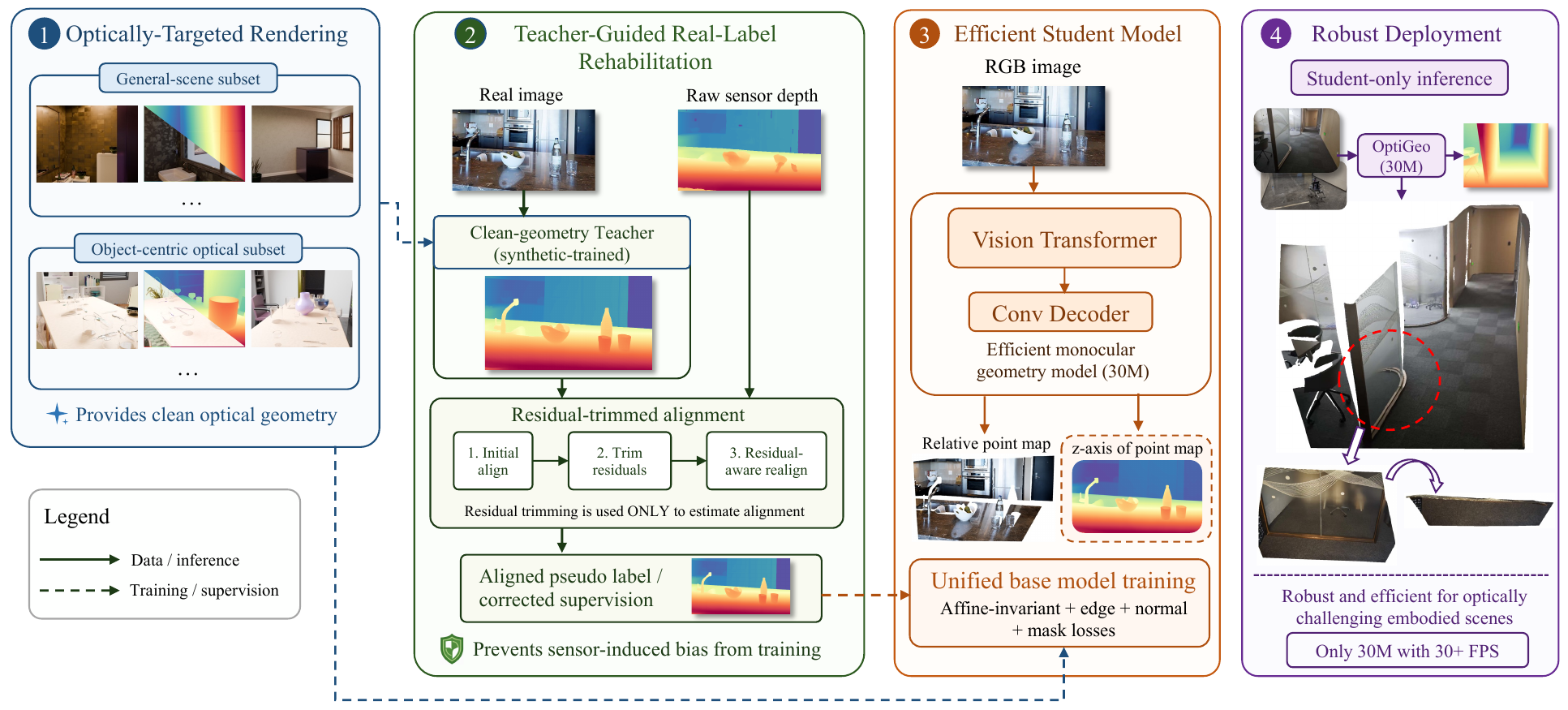}
    \vspace{-14pt}
    \caption{
    OptiGeo uses optically-targeted rendering and teacher-guided real-label rehabilitation to correct sensor-induced supervision bias during base-model training.
    Clean rendered geometry and corrected real labels jointly supervise an efficient 30M-parameter student with a unified base model training pipeline.
    At inference, only the student is deployed for robust and efficient geometry in optically challenging scenes.
    }
    \label{fig:optigeo_overview}
    \vspace{-14pt}
\end{figure}

The empirical study suggests that optical robustness should be addressed during base-model training, rather than through post-hoc transparent-scene specialization. 
OptiGeo (as shown in Fig.~\ref{fig:optigeo_overview}) follows this principle through three components: correcting biased real supervision with teacher-guided label rehabilitation (Sec.~\ref{sec:teacher_refine}), injecting clean optical geometry with transparency-targeted rendering (Sec.~\ref{sec:target_rendering}), and training a compact deployable student with a unified point-map objective (Sec.~\ref{sec:training_objective}).

Given an input image $\mathbf I$, the student model $f_{\theta}$ predicts a dense relative point map $\hat{\mathbf P}=f_{\theta}(\mathbf I)$, with depth obtained from the $z$-axis component $\hat{\mathbf D}=\Pi_z(\hat{\mathbf P})$. 
During training, a clean-geometry teacher $f_{\theta^t}$ predicts $\hat{\mathbf P}^{t}=f_{\theta^t}(\mathbf I)$ and $\hat{\mathbf D}^{t}=\Pi_z(\hat{\mathbf P}^{t})$ to construct corrected supervision. 
The teacher is used only for training; inference uses the compact student alone.

\vspace{-4pt}
\subsection{Teacher-Guided Real-Label Rehabilitation}
\vspace{-4pt}
\label{sec:teacher_refine}

Real depth labels provide metric scale and real-world appearance, but can be locally biased in optically challenging regions. OptiGeo preserves their reliable calibration cues while replacing corrupted geometry with a synthetic-trained clean-geometry teacher. 

\vspace{0.03in}\noindent \textbf{Clean-geometry teacher training.}
Before rehabilitating real labels, we first train a clean-geometry teacher using only synthetic samples with reliable geometric supervision.
The teacher follows the same relative point-map formulation as the student (Sec.~\ref{sec:training_objective}), but adopts a larger geometry model to provide more stable full-image predictions.

Given a real training sample with image $\mathbf I$, sensor depth $\mathbf D^s$, and valid mask $\mathbf M^s$, the frozen teacher predicts a relative point map $\hat{\mathbf P}^{t}=f_{\theta^t}(\mathbf I)$, and we take its $z$-axis component as the teacher depth, $\hat{\mathbf D}^{t}=\Pi_z(\hat{\mathbf P}^{t})$.
Since $\hat{\mathbf D}^{t}$ is relative, we align it to the sensor depth with residual-trimmed scale-shift fitting. 
Let $p$ denote a pixel location and $\Omega_s=\{p\mid M^s_p=1,\; D^s_p,\hat D^t_p \text{ are finite}\}$ be the initial valid fitting set. 
We first estimate an initial least-squares alignment and compute its residual:
\begin{equation}
    (a_0,b_0)=
    \arg\min_{a,b}
    \sum_{p\in\Omega_s}
    \left(a\hat{D}^{t}_{p}+b-D^{s}_{p}\right)^2,
    \qquad
    r^{(0)}_{p}=
    \left|a_0\hat{D}^{t}_{p}+b_0-D^{s}_{p}\right|.
\end{equation}
\vspace{-6pt}

Large residuals indicate strong disagreement between the clean-geometry teacher and the sensor label, often caused by background leakage on glass, distorted reflections, or locally inconsistent specular measurements. 
We remove the top $\tau$ fraction of high-residual pixels and obtain the residual-trimmed fitting set 
$\Omega_r=\{p\in\Omega_s\mid r^{(0)}_{p}\leq Q_{1-\tau}(r^{(0)})\}$, where $Q_{1-\tau}(\cdot)$ denotes the $(1-\tau)$ quantile over $\Omega_s$. 
Using the retained pixels, we recompute the final least-squares scale and shift and apply it to the full teacher prediction:
\begin{equation}
    (a,b)=
    \arg\min_{a,b}
    \sum_{p\in\Omega_r}
    \left(a\hat{D}^{t}_{p}+b-D^{s}_{p}\right)^2,
    \qquad
    \tilde{\mathbf D}^{t}=a\hat{\mathbf D}^{t}+b.
\end{equation}
\vspace{-6pt}

Importantly, $\Omega_r$ is used only to estimate the global alignment parameters; it does not restrict supervision to a subset of pixels. 
The aligned teacher depth $\tilde{\mathbf D}^{t}$ serves as full-image corrected supervision, with metric scale calibrated by reliable real pixels and dense geometry supplied by the synthetic-trained teacher.

\vspace{-6pt}
\subsection{Targeted Transparency Data Rendering}
\vspace{-4pt}
\label{sec:target_rendering}

The goal of rendering in OptiGeo is not to build a large transparency-specific domain for post-hoc fine-tuning. 
Instead, we use a compact targeted rendering set to provide clean optical geometry that real sensors cannot reliably annotate. 
This follows our empirical observation that optical failures mainly appear as local geometry distortions around transparent objects, reflective surfaces, and optical boundaries.

We build the rendering pipeline on Infinigen~\cite{raistrick2023infinite}, which allows controllable scene geometry, materials, lighting, and cameras. 
We render two complementary subsets. 
The general-scene subset maintains broad geometric diversity and prevents the model from over-specializing to tabletop transparent scenes. 
The object-centric optical subset focuses on glassware and reflective objects, providing clean depth and point-map supervision for regions where RGB-D and ToF labels are often missing, biased, or background-leaking.

For the object-centric subset, each scene is initialized as an indoor tabletop environment and populated with randomized transparent objects, including $3$--$6$ cups and $3$--$6$ wineglasses. 
This object density exposes the model to transparent-object boundaries, inter-object occlusions, and local geometry interactions, while remaining sufficiently sparse for unambiguous supervision. 
We enforce a minimum object spacing of about $0.12$m to avoid severe overlap, and add local lighting to make transparent boundaries more visible. 
For each scene, we sample eight orbiting views with camera distances from $1.2$ to $2.0$m, providing diverse viewpoints while keeping the target objects prominent.

\vspace{-6pt}
\subsection{Student Training Objective}
\vspace{-6pt}
\label{sec:training_objective}

We train the compact student with an affine-invariant point-map objective following MoGe-style~\cite{wang2025moge} relative geometry training. 
Given an input image $\mathbf I$, the student predicts a dense relative point map $\hat{\mathbf P}\in\mathbb{R}^{H\times W\times 3}$ and a reliability mask $\hat{\mathbf M}\in[0,1]^{H\times W}$, with the evaluation depth obtained from the $z$-axis component of $\hat{\mathbf P}$.

As summarized in Table~\ref{datasets}, training uses a curated 9M-sample corpus from 21 datasets spanning diverse domains. 
For standard samples, we use the filtered valid geometric labels. 
For rehabilitated real samples, the biased raw sensor depth is replaced by the aligned teacher depth $\tilde{\mathbf D}^{t}$ from Sec.~\ref{sec:teacher_refine}, which is back-projected into point-map supervision. 

The relative geometry loss combines global and local affine alignment:
\begin{equation}
    \mathcal L_{\mathrm{rel}}
    =
    \mathcal L_{\mathrm{global}}
    +
    \sum_{\gamma\in\mathcal A_t}
    \mathcal L_{S(\gamma)},
\end{equation}
where $\mathcal L_{\mathrm{global}}$ aligns the prediction and target under global scale--shift ambiguity, and $\mathcal L_{S(\gamma)}$ applies patch-level affine alignment at multiple scales to preserve local structures and sharp boundaries.

We further use geometry-aware auxiliary losses:
\begin{equation}
    \mathcal L
    =
    \mathcal L_{\mathrm{rel}}
    +
    \lambda_{\mathrm n}\mathcal L_{\mathrm{normal}}
    +
    \lambda_{\mathrm e}\mathcal L_{\mathrm{edge}}
    +
    \lambda_{\mathrm m}\mathcal L_{\mathrm{mask}}.
\end{equation}
In this unified objective, transparency-targeted samples provide clean synthetic optical geometry, and rehabilitated real samples provide corrected supervision without directly inheriting sensor-induced bias.

%% file: sec/5_experiments.tex
\vspace{-8pt}
\section{Experiments}
\vspace{-8pt}

\begin{table}[t]
    \caption{Summary of datasets used for training.}
    \label{datasets}
    \centering
    \scriptsize
    \renewcommand{\arraystretch}{1.10}
    \setlength{\tabcolsep}{2.5pt}

    \begin{minipage}[t]{0.46\textwidth}
        \centering
        \resizebox{\linewidth}{!}{
        \begin{tabular}{l c c c}
            \specialrule{0.12em}{0em}{0em}
            Name & Domain & \# Frames & Syn. \\
            \hline
            ARKitScenes~\cite{baruch1arkitscenes} & Indoor & $441$K & N \\
            BlendedMVS~\cite{yao2020blendedmvs} & In-the-wild & $109$K & N \\
            Taskonomy~\cite{zamir2018taskonomy} & Indoor & $4.6$M & N \\
            Scannet++~\cite{yeshwanth2023scannet++} & Indoor & $398$K & N \\
            STD~\cite{dai2022domain} & Indoor & $4.5$K & N \\
            Waymo~\cite{sun2020scalability} & Outdoor/Driving & $790$K & N \\
            OptiGeo & Indoor & $7$K & Y \\
            FSD~\cite{wen2025stereo} & Outdoor/In-the-wild & $1.04$M & Y \\
            Hypersim~\cite{roberts2021hypersim} & Indoor & $64$K & Y \\
            IRS~\cite{wang2019irs} & Indoor & $94$K & Y \\
            KenBurns~\cite{niklaus20193d} & In-the-wild & $72$K & Y \\
            \specialrule{0.12em}{0em}{0em}
        \end{tabular}}
    \end{minipage}
    \hfill
    \begin{minipage}[t]{0.49\textwidth}
        \centering
        \resizebox{\linewidth}{!}{
        \begin{tabular}{l c c c}
            \specialrule{0.12em}{0em}{0em}
            Name & Domain & \# Frames & Syn. \\
            \hline
            MatrixCity~\cite{li2023matrixcity} & Outdoor/Driving & $354$K & Y \\
            MidAir~\cite{fonder2019mid} & Outdoor/In-the-wild & $423$K & Y \\
            MVS-Synth~\cite{huang2018deepmvs} & Outdoor/Driving & $12$K & Y \\
            Spring~\cite{mehl2023spring} & In-the-wild & $5$K & Y \\
            Structured3D~\cite{zheng2020structured3d} & Indoor & $76$K & Y \\
            TartanAir~\cite{wang2020tartanair} & In-the-wild & $259$K & Y \\
            PointOdyssey~\cite{zheng2023pointodyssey} & Indoor & $79$K & Y \\
            UrbanSyn~\cite{gomez2025all} & Outdoor/Driving & $7$K & Y \\
            Dynamic-Replica~\cite{karaev2023dynamicstereo} & Indoor & $143$K & Y \\
            FGD & Indoor/Outdoor & $23$K & Y \\
            \hline
            \textbf{Total} & & \textbf{9M} & \\
            \specialrule{0.12em}{0em}{0em}
        \end{tabular}}
    \end{minipage}

    \vspace{-12pt}
\end{table}

\textbf{Implementation Details.} OptiGeo uses an ViT-Small~\cite{dosovitskiy2021imageworth16x16words} encoder pre-trained with DINOv3~\cite{simeoni2025dinov3}. During training, the encoder and decoder are trained with initial learning rates of $1\times 10^{-5}$ and $1\times 10^{-4}$, respectively, and the learning rate is halved every 20K iterations. The full model is trained for 100K iterations using 32 NVIDIA H20 GPUs. 

\vspace{0.03in} \noindent \textbf{Datasets.}
As shown in Table~\ref{datasets}, OptiGeo is trained on 21 datasets comprising approximately 9 million frames.
We evaluate transparent-scene geometry on ClearGrasp Real~\cite{sajjan2020clear} and TransCG~\cite{fang2022transcg}, where we use first 3 test Scenes 7, 8, and 11, comprising 707 frames in total; Scene 5 is excluded because it contains almost no transparent objects.
For general zero-shot depth evaluation, we use NYUv2~\cite{silberman2012indoor}, ETH3D~\cite{schops2019bad}, iBims-1~\cite{koch2018evaluation,koch2020comparison}, Sintel~\cite{butler2012naturalistic}, and DIODE~\cite{vasiljevic2019diode}.
All evaluation datasets are excluded from training.

\textbf{Evaluation Metrics.}
We measure depth accuracy using AbsRel, SiLog, RMSE, MAE, and threshold accuracy $\delta_t$, and evaluate depth quality using boundary F1.
AbsRel is computed as $\frac{1}{N}\sum_i |\hat d_i-d_i|/d_i$.
RMSE is $\sqrt{\frac{1}{N}\sum_i(\hat d_i-d_i)^2}$, and MAE is $\frac{1}{N}\sum_i|\hat d_i-d_i|$.
SiLog is computed as $\sqrt{\frac{1}{N}\sum_i g_i^2-(\frac{1}{N}\sum_i g_i)^2}$, where $g_i=\log \hat d_i-\log d_i$.
The threshold accuracy $\delta_t$ measures the percentage of pixels satisfying $\max(\hat d_i/d_i,d_i/\hat d_i)<t$.
For transparent-scene evaluation, we report depth errors and use $\delta_{1.10}$.
For boundary sharpness, we report boundary F1 following Depth Pro~\cite{bochkovskiydepth}.

\vspace{0.03in} \noindent \textbf{Baselines.}
For transparent-scene evaluation, we compare OptiGeo on ClearGrasp Real and TransCG against two groups of baselines: geometry foundation models~\cite{ke2025marigold,hu2024metric3d,piccinelli2025unidepthv2,lin2025depth,wang2025vggt,wang2025moge,wang2025moge2} and transparency-targeted methods~\cite{xu2025diffusion,liu2025monocular}. 
For general zero-shot relative-depth evaluation, we further test on~\cite{silberman2012indoor,schops2019bad,koch2018evaluation,koch2020comparison,vasiljevic2019diode} and compare with representative monocular and multi-view geometry foundation models. All predictions are aligned to ground truth with a single global scale and shift before computing relative-depth metrics.

\vspace{-6pt}
\subsection{Main Results}
\vspace{-4pt}

We assess the zero-shot performance of OptiGeo and compare it to several state-of-the-art methods on monocular affine-invariant depth estimation and boundary sharpness. 

\vspace{0.03in}\noindent \textbf{Quantitative results on transparent-scene benchmarks:}
Tables~\ref{tab:cleargrasp_all} and~\ref{tab:transcg} evaluate OptiGeo on ClearGrasp Real and TransCG, respectively.
With only 30M parameters, OptiGeo achieves strong performance on both benchmarks, reaching 0.019 AbsRel and 97.92\% $\delta_{1.10}$ on ClearGrasp Real, and 4.36 AbsRel and 89.55\% $\delta_{1.10}$ on TransCG.
These results demonstrate a strong accuracy--efficiency trade-off and consistent generalization across real transparent-object scenes.
As shown in Table~\ref{tab:scaleup_comparison}, scaling the model further improves performance on both benchmarks.

\vspace{0.03in}\noindent \textbf{Quantitative results on general zero-shot depth estimation:}
Table~\ref{table:relative_depth} evaluates zero-shot relative-depth performance on NYUv2, ETH3D, iBims-1, Sintel, and DIODE after global scale-and-shift alignment.
Although OptiGeo is specifically designed to improve geometry in optically challenging scenes, it maintains strong performance across general indoor, outdoor, and synthetic benchmarks.
The per-dataset results further show that OptiGeo performs particularly well on Sintel, where complex appearance, motion, and non-Lambertian effects make monocular geometry estimation challenging.
Overall, these results show that the proposed bias-aware optical training does not over-specialize the model to transparent-object scenarios, but preserves broad zero-shot geometry generalization.

\vspace{0.03in}\noindent \textbf{Effect of model scaling.}
As shown in Table~\ref{tab:scaleup_comparison}, increasing model capacity yields consistent gains on ClearGrasp Real, TransCG, and general zero-shot depth evaluation.
The Hplus variant reaches 0.012 AbsRel on ClearGrasp Real, 3.64 on TransCG, and 4.56 on general relative-depth benchmarks.
This consistent scaling behavior suggests that the gains arise from the proposed training strategy rather than a specific model size, while the 30M variant remains the preferred efficiency-oriented deployment setting.

\begin{table}[t]
\centering
\begin{minipage}[t]{0.64\linewidth}
\centering
\vspace{0pt}
\caption{\textbf{ClearGrasp Real.} Comparison on the real-world test set over the full image. RMSE and MAE are reported in mm, and $\delta$ is reported in percentage.}
\label{tab:cleargrasp_all}
\tiny
\renewcommand{\arraystretch}{1.0}
\resizebox{\linewidth}{!}{%
\begin{tabular}{lcccccc}
\toprule
\textbf{Method}
& \textbf{\#Params}
& AbsRel$\downarrow$
& SiLog$\downarrow$
& RMSE$\downarrow$
& MAE$\downarrow$
& $\delta_{1.10}\uparrow$ \\
\midrule

DKT~\cite{xu2025diffusion}
& 1300M & 0.077 & 0.098 & 72.89 & 53.09 & 72.75 \\

Marigold~\cite{ke2025marigold}
& 948M & 0.049 & 0.062 & 48.64 & 31.76 & 87.07 \\

Metric3D V2~\cite{hu2024metric3d}
& 412M & \textbf{0.019} & 0.032 & \underline{21.12} & \textbf{12.66} & \underline{97.58} \\

UniDepth V2~\cite{piccinelli2025unidepthv2}
& 400M & 0.021 & \underline{0.030} & 21.25 & 14.34 & 97.22 \\

DAv3-Large~\cite{lin2025depth}
& 350M & 0.029 & 0.045 & 40.56 & 22.03 & 93.08 \\

MODEST~\cite{liu2025monocular}
& 120M & 0.106 & 0.138 & 112.1
& 73.11 & 58.56 \\

VGGT~\cite{wang2025vggt}
& 1200M & 0.020 & 0.032 & 23.52 & 13.96 & 96.24 \\

MoGe~\cite{wang2025moge}
& 314M & \textbf{0.019} & 0.036 & 24.00 & \underline{12.74} & 96.45 \\

MoGe-2~\cite{wang2025moge2}
& 326M & \textbf{0.019} & 0.031 & 21.95
& {12.95} & 97.08 \\

\midrule
\rowcolor{gray!12}
OptiGeo
& \textbf{30M} & \textbf{0.019} & \textbf{0.029} & \textbf{20.26} & 13.01 & \textbf{97.92} \\

\bottomrule
\end{tabular}%
}
\end{minipage}%
\hfill%
\begin{minipage}[t]{0.33\linewidth}
\centering
\vspace{0pt}
\caption{\textbf{TransCG first 3 test scenes.} Extended transparent benchmark.}
\label{tab:transcg}

\tiny
\setlength{\tabcolsep}{2.5pt}
\renewcommand{\arraystretch}{1.05}

\resizebox{\linewidth}{!}{%
\begin{tabular}{lccc}
\toprule
\textbf{Method}
& \textbf{\#Params}
& AbsRel$\downarrow$
& $\delta_{1.10}\uparrow$ \\
\midrule

DKT
& 1300M
& 6.81
& 74.81 \\

Metric3D V2
& 412M
& 4.64
& 87.50 \\

DAv3-Large
& 350M
& 5.45
& 82.01 \\

UniDepth V2
& 350M
& 7.98
& 69.58 \\

MODEST
& 120M
& 10.5
& 56.04 \\

MoGe
& 314M
& \textbf{4.23}
& 88.37 \\

MoGe-2
& 326M
& 4.64
& 86.75 \\

VGGT
& 1200M
& 4.39
& \underline{89.42} \\

\midrule

\rowcolor{gray!12}
OptiGeo
& \textbf{30M}
& \underline{4.36}
& \textbf{89.55} \\

\bottomrule
\end{tabular}%
}
\end{minipage}
\vspace{-8pt}
\end{table}

\begin{table}[t]
  \caption{
  Quantitative results for relative depth estimation.
  \textit{AbsRel} and $\delta_1$ are in percentage.
  The best values are highlighted in \textbf{bold}, and the second-best ones are \underline{underlined}.
  \textcolor{gray!50}{Gray numbers} denote models trained on respective benchmarks.
  }
  \vspace{-4pt}
  \label{table:relative_depth}
  \centering
  \setlength{\tabcolsep}{2.0pt}
  \renewcommand{\arraystretch}{1.15}
  \scriptsize
  \resizebox{\textwidth}{!}{%
  \begin{tabular}{l c | c c | c c | c c | c c | c c | c c}
    \toprule
    \multirow{2}{*}{Method}
      & \multirow{2}{*}{\#Params}
      & \multicolumn{2}{c|}{NYUv2}
      & \multicolumn{2}{c|}{ETH3D}
      & \multicolumn{2}{c|}{iBims-1}
      & \multicolumn{2}{c|}{Sintel}
      & \multicolumn{2}{c|}{DIODE}
      & \multicolumn{2}{c}{Average} \\
      &
      & AbsRel$\downarrow$ & $\delta_1\uparrow$
      & AbsRel$\downarrow$ & $\delta_1\uparrow$
      & AbsRel$\downarrow$ & $\delta_1\uparrow$
      & AbsRel$\downarrow$ & $\delta_1\uparrow$
      & AbsRel$\downarrow$ & $\delta_1\uparrow$
      & AbsRel$\downarrow$ & $\delta_1\uparrow$ \\
    \midrule

    ZoeDepth~\cite{bhat2023zoedepth}
      & 345M
      & \textcolor{gray!50}{4.76} & \textcolor{gray!50}{97.3}
      & 7.27 & 94.2
      & 5.85 & 95.7
      & 21.8 & 69.2
      & 7.80 & 90.9
      & 9.50 & 89.5 \\

    PPD~\cite{xupixel}
      & 500M
      & 5.16 & 97.0
      & 7.98 & 91.1
      & 5.34 & 95.9
      & 18.8 & 74.4
      & 7.68 & 90.5
      & 8.99 & 89.8 \\

    MASt3R~\cite{murai2025mast3r}
      & $\sim$700M
      & 4.67 & 96.7
      & 4.64 & 97.0
      & 4.62 & 95.6
      & 21.3 & 70.3
      & 5.79 & 94.1
      & 8.20 & 90.7 \\

      Marigold~\cite{ke2025marigold}
      & $\sim$950M
      & 4.63 & 97.3
      & 6.08 & 96.3
      & 4.35 & 97.2
      & 21.2 & 75.0
      & 6.34 & 94.3
      & 8.52 & 92.0 \\

    GeoWizard~\cite{fu2024geowizard}
      & $\sim$900M 
      & 4.69 & 97.4
      & 6.90 & 93.9
      & 4.50 & 97.1
      & 17.8 & 76.2
      & 7.03 & 92.7
      & 8.18 & 91.5 \\

    Metric3D V2~\cite{hu2024metric3d}
      & 412M
      & 3.94 & 97.6
      & \textbf{3.24} & \textbf{99.0}
      & \underline{3.28} & \underline{98.3}
      & 26.6 & 71.7
      & \textbf{2.75} & \textbf{98.7}
      & 7.96 & \underline{93.1} \\

    VGGT~\cite{wang2025vggt}
      & 1200M
      & \textbf{3.01} & \textbf{98.4}
      & \underline{3.47} & \underline{97.7}
      & 3.64 & 96.9
      & 18.1 & 76.2
      & 5.15 & 94.6
      & \underline{6.67} & 92.8 \\

    DA V1~\cite{yang2024depth}
      & 335M
      & \textcolor{gray!50}{3.82} & \textcolor{gray!50}{98.3}
      & 6.23 & 95.2
      & 4.23 & 97.3
      & 20.1 & 71.8
      & 6.75 & 92.6
      & 8.23 & 91.0 \\

    DA V2~\cite{yang2024depth2}
      & 335M
      & 4.16 & \underline{97.9}
      & 4.63 & 97.2
      & 3.44 & \underline{98.3}
      & \underline{17.1} & \underline{76.6}
      & 5.41 & 94.6
      & 6.95 & 92.9 \\

    DA V3~\cite{lin2025depth}
      & 350M
      & \underline{3.39} & \textbf{98.4}
      & 4.37 & 96.9
      & \textbf{2.96} & \textbf{98.6}
      & 17.4 & 75.8
      & \underline{4.85} & \underline{95.5}
      & \textbf{6.59} & 93.0 \\

    \rowcolor{gray!12}
    OptiGeo
      & \textbf{30M}
      & 4.41 & 97.5
      & 4.89 & 97.3
      & 3.99 & 97.4
      & \textbf{15.5} & \textbf{78.5}
      & 4.93 & \underline{95.5}
      & 6.74 & \textbf{93.2} \\

    \bottomrule
  \end{tabular}%
  }
  \vspace{-8pt}
\end{table}

\begin{table}[t]
\centering

% ============================================================
% Left: Scale-up Comparison
% ============================================================
\begin{minipage}[t]{0.48\linewidth}
\centering
\vspace{0pt}

\caption{Scale-up comparison of OptiGeo.}
\label{tab:scaleup_comparison}

\tiny
\setlength{\tabcolsep}{2.5pt}
\renewcommand{\arraystretch}{1.00}

\resizebox{0.68\linewidth}{!}{%
\begin{tabular}{lccc}
\toprule
\textbf{Method}
& \textbf{\#Params}
& AbsRel$\downarrow$
& $\delta\uparrow$ \\
\midrule

\multicolumn{4}{c}{\textbf{ClearGrasp Real}} \\
\midrule

OptiGeo
& \textbf{30M}
& 0.019
& 97.92 \\

OptiGeo-Large
& 314M
& {0.015}
& {98.80} \\

OptiGeo-Hplus
& 940M
& \textbf{0.012}
& \textbf{98.93} \\

\midrule
\multicolumn{4}{c}{\textbf{TransCG First Three Test Scenes}} \\
\midrule

OptiGeo
& \textbf{30M}
& 4.36
& 89.55 \\

OptiGeo-Large
& 314M
& 3.74
& {93.29} \\

OptiGeo-Hplus
& 940M
& \textbf{3.64}
& \textbf{93.97} \\

\midrule
\multicolumn{4}{c}{\textbf{General Relative-Depth Comparison}} \\
\midrule

OptiGeo
& \textbf{30M}
& 6.74
& 93.2 \\

OptiGeo-Large
& 314M
& 4.99
& {95.4} \\

OptiGeo-Hplus
& 940M
& \textbf{4.56}
& \textbf{96.0} \\

\bottomrule
\end{tabular}%
}

\end{minipage}%
\hfill%
% ============================================================
% Right: Edge F1
% ============================================================
\begin{minipage}[t]{0.48\linewidth}
\centering
\vspace{0pt}

\caption{\textbf{Boundary sharpness.}
Evaluation using F1 scores ($\uparrow$) in percentages.}
\label{tab:qualitative_comparison_boundary}

\tiny
\renewcommand{\arraystretch}{1.12}

\resizebox{\linewidth}{!}{%
\begin{tabular}{lccc}
\toprule
\textbf{Method}
& \textbf{\#Params}
& iBims-1
& HAMMER \\
\midrule

ZoeDepth~\cite{bhat2023zoedepth}
& 345M & 2.47 & 0.17 \\

DA V1~\cite{yang2024depth}
& 335M & 3.68 & 0.76 \\

DA V2~\cite{yang2024depth2}
& 335M & 13.9 & 4.74 \\

DA V3-Large~\cite{lin2025depth}
& 350M & 11.8 & 2.19 \\

UniDepth V1~\cite{piccinelli2024unidepth}
& 347M & 2.35 & 0.06  \\

UniDepth V2~\cite{piccinelli2025unidepthv2}
& 354M & 11.2 & 4.40  \\

Metric3D V2~\cite{hu2024metric3d}
& 412M & 7.36 & 1.40 \\

MASt3R~\cite{murai2025mast3r}
& $\sim$700M & 1.24 & 0.05 \\

Depth Pro~\cite{bochkovskiydepth}
& 504M & 14.3 & \textbf{5.36} \\

MoGe~\cite{wang2025moge}
& 314M & 11.4 & 3.89 \\

\midrule
\rowcolor{gray!12}
OptiGeo
& \textbf{30M}
& \textbf{15.5}
& 4.30 \\

\bottomrule
\end{tabular}%
}

\end{minipage}

\vspace{-10pt}
\end{table}

\vspace{0.04in}\noindent\textbf{Quantitative measurements for boundary sharpness.}
We evaluate boundary sharpness using boundary F1~\cite{bochkovskiydepth} on iBims-1~\cite{koch2020comparison,koch2018evaluation} and HAMMER~\cite{jung2023importance}.
As shown in Table~\ref{tab:qualitative_comparison_boundary}, OptiGeo achieves the best F1 score on iBims-1 and remains competitive on HAMMER, outperforming most large-scale baselines with only 30M parameters.
This indicates that OptiGeo not only improves depth accuracy, but also preserves sharp geometric discontinuities, which are important for transparent-object boundaries, reflective surfaces, and obstacle reasoning in embodied scenes.

\vspace{-8pt}
\subsection{Evaluation in Real Scenes}
\vspace{-6pt}

% \begin{figure}[t]
%     \centering
%     \includegraphics[width=\textwidth]{figures/real.pdf}
%     \caption{Real-world optically challenging scenes where transparent and reflective objects cause unreliable depth measurements.}
%     \label{fig:real_scene}
% \end{figure}

Beyond standard benchmarks, we evaluate OptiGeo in real optically challenging scenes to examine its practicality for robot perception.

\begin{wrapfigure}{r}{0.48\textwidth}
\centering
\vspace{-10pt}
\includegraphics[width=0.99\linewidth]{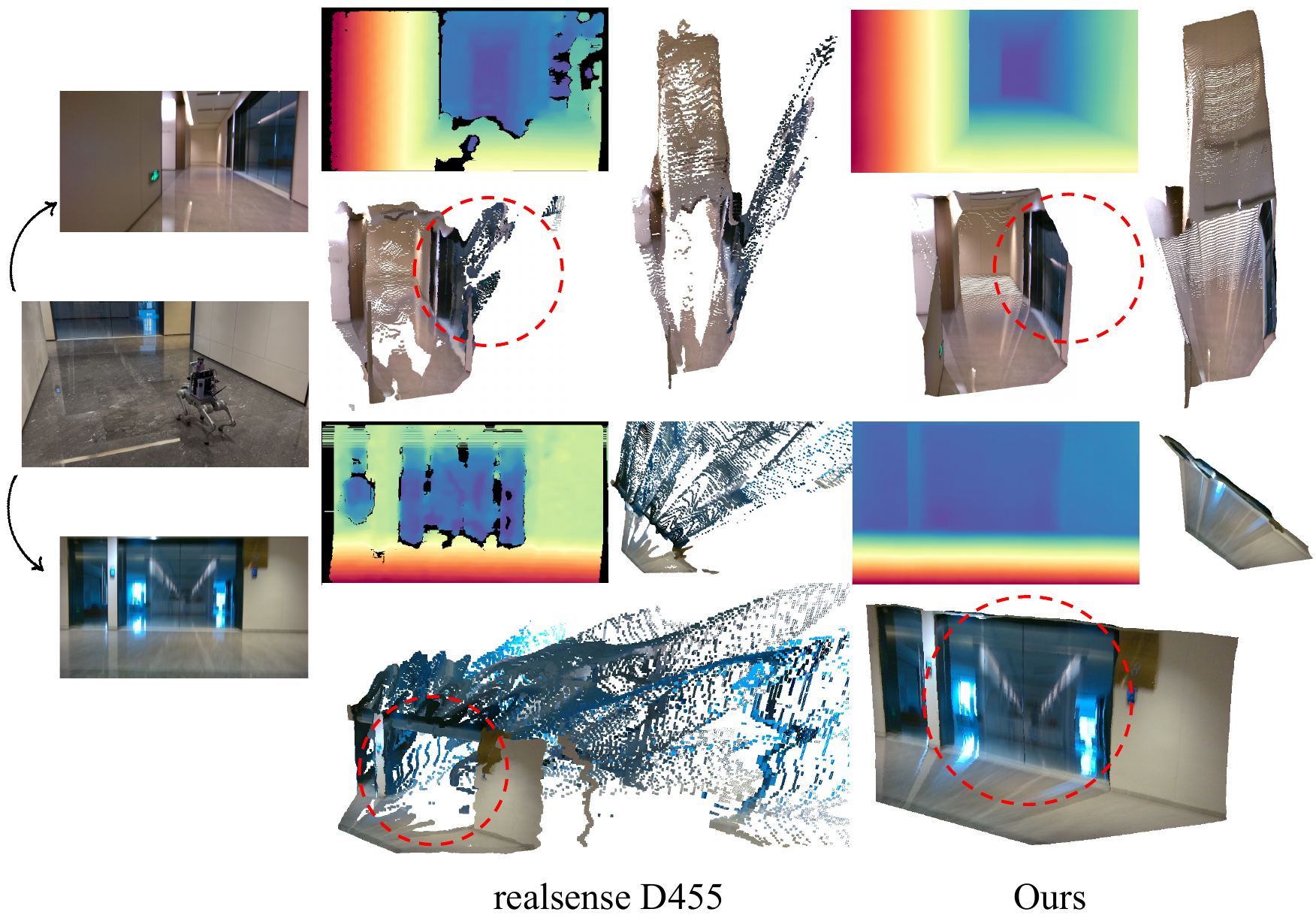}
\vspace{-14pt}
\caption{Glass-door scenes from side and frontal views. Real sensor largely fails on the transparent door, while OptiGeo recovers coherent planar geometry.}
\label{fig:real_glass_door}
\vspace{-10pt}
\end{wrapfigure}
As shown in Fig.~\ref{fig:real_glass_door}, RealSense D455 produces missing, fragmented, or background-shifted depth on glass doors, which can cause incorrect free-space reasoning for navigation.
OptiGeo instead recovers a coherent planar structure from both side and frontal views, and remains stable even under image shake, while preserving fine details around door boundaries.
Fig.~\ref{fig:real_transparent_object} also shows scenes with transparent covers and surrounding glass structures.
RealSense D455 largely fails in these regions due to penetration, reflection, and sparse returns, whereas OptiGeo produces dense and coherent geometry for both the scene layout and transparent targets.
These results indicate that OptiGeo learns useful optical-region geometry without transparent-object masks, multi-view fusion, or test-time refinement, supporting its use as an efficient feed-forward geometry module for embodied perception.

\noindent\textbf{Quantitative real-robot evaluation.}
To further evaluate its practical utility, we conduct closed-loop navigation experiments on a Unitree Go2 Air equipped with a RealSense D435. We use a fine-tuned metric variant of OptiGeo, and convert its decoder to TensorRT for accelerated edge deployment.
Transparent obstacles are not pre-encoded in the global map, and obstacle avoidance relies entirely on online depth point clouds projected onto the Nav2~\cite{macenski2020marathon2} local costmap.
All navigation settings are kept identical except for the depth source, i.e., RealSense D435 or OptiGeo.
We evaluate three transparent-object scenes with 20 repeated trials per scene.
A trial is considered successful if the robot reaches the goal without collision or manual intervention.
As shown in Table~\ref{tab:robot_success}, OptiGeo substantially improves the average navigation success rate from 10.0\% to 78.3\%, demonstrating that its recovered geometry can directly support downstream navigation in optically challenging environments.

\vspace{-6pt}

\begin{table}[h]
\centering
\caption{\textbf{Real-robot task evaluation.}
Navigation success rate under three transparent-object scenarios.}
\label{tab:robot_success}

\scriptsize
\setlength{\tabcolsep}{5pt}
\renewcommand{\arraystretch}{1.05}

\begin{tabular}{lcccc}
\toprule
\textbf{Method}
& \textbf{Scene 1}
& \textbf{Scene 2}
& \textbf{Scene 3}
& \textbf{Avg.} \\
\midrule

RealSense D435
& 5\%
& 15\%
& 10\%
& 10.0\% \\

\rowcolor{gray!12}
\textbf{OptiGeo}
& 70\%
& 85\%
& 80\%
& \textbf{78.3\%} \\

\bottomrule
\end{tabular}

\vspace{-10pt}
\end{table}

%% file: sec/6_conclusion.tex
\vspace{-13pt}

\section{Conclusion}
\vspace{-8pt}

We presented OptiGeo, an efficient monocular geometry framework for embodied perception in optically challenging scenes.
Motivated by our empirical study, we treat transparent and reflective regions as localized failures within general geometry training, where biased real sensor labels can cause models to inherit sensor failure patterns.
OptiGeo addresses this problem by rehabilitating real labels with a clean-geometry teacher, using residual-trimmed alignment to suppress biased sensor regions, and injecting compact transparency-targeted rendering as clean optical supervision.
These corrected real labels and rendered samples jointly train a 30M feed-forward student.
Experiments on ClearGrasp Real, general zero-shot depth benchmarks and real optically challenging scenes show that OptiGeo improves transparent-region geometry while preserving broad generalization and efficient deployment.

\vspace{-12pt}
\section{Limitations}
\vspace{-8pt}

OptiGeo is a monocular model and therefore does not explicitly enforce multi-frame temporal consistency~\cite{lyu2026streamingdepth}.
Since it is built on a monocular relative-geometry base model, metric deployment still requires external finetuning or scale calibration~\cite{liu2026foundationgeo}.
Our model may fail in visually ambiguous optical regions with weak semantic or texture priors, such as large glass doors and windows.

\vspace{-8pt}
\section{Acknowledgment}
\vspace{-8pt}
We would like to thank Yuduo Wang and Changyan Li for their support in real-robot experiments.
The research was supported by the Hong Kong Research Grants Council (RGC) through the General Research Fund (Grants No. 17202422, 17212923, and 17215025), the Theme-based Research Scheme (Grant No. T45-701/22-R), and the Strategic Topics Grant (Grant No. STG3/E-605/25-N). Additionally, part of this research was conducted at the JC STEM Lab of Robotics for Soft Materials, funded by The Hong Kong Jockey Club Charities Trust.

%% file: sec/X_suppl.tex
\clearpage
\appendix
\setcounter{page}{1}
\section*{Supplementary Material}
\addcontentsline{toc}{section}{Supplementary Material}

\section{Ablation Study}
\label{sec:ablation}

We conduct ablation studies to analyze the two key components of OptiGeo: teacher-guided real-label rehabilitation with residual-trimmed alignment, and transparency-targeted rendering. The first group of ablations isolates different real-label refinement strategies under a compact training setting. The second group evaluates the effect of adding our targeted rendering set under the full-data training setting.

\subsection{Effect of Real-label Rehabilitation}

\paragraph{Setup and variants.}
To make the comparison efficient and focused, we train all variants in this group for 25K iterations using 8 NVIDIA H20 GPUs. The training subset contains only ARKitScenes~\cite{baruch1arkitscenes} and Hypersim~\cite{roberts2021hypersim}, where ARKitScenes provides real RGB-D supervision and Hypersim provides clean synthetic geometry. This compact setting allows us to directly analyze how different refinement strategies affect biased real supervision while maintaining a controlled real-synthetic data mixture.

We compare three variants. \textbf{Base} directly uses the original supervision without teacher-guided label rehabilitation. \textbf{+ Refine} replaces real sensor labels with teacher predictions aligned to sensor depth using a single global scale and shift estimated over all valid sensor pixels. \textbf{+ Trim + Refine} adopts the proposed residual-trimmed alignment, where high-residual pixels are excluded only when estimating the global scale and shift, while the final aligned teacher prediction is still used as dense supervision. Therefore, the comparison between \textbf{+ Refine} and \textbf{+ Trim + Refine} directly evaluates whether suppressing unreliable sensor pixels during alignment improves real-label rehabilitation.

\begin{table}[h]
\centering
\begin{minipage}[t]{0.68\linewidth}
\centering
\caption{\textbf{ClearGrasp Real.} Ablation study on the real-world test set using affine-invariant depth metrics. RMSE and MAE are reported in mm.}
\label{tab:cleargrasp_ablation}
\tiny
\renewcommand{\arraystretch}{1.0}
\resizebox{\linewidth}{!}{%
\begin{tabular}{lccccc}
\toprule
\textbf{Variant}
& AbsRel$\downarrow$
& SiLog$\downarrow$
& RMSE$\downarrow$
& MAE$\downarrow$
& $\delta_{1}\uparrow$ \\
\midrule
Base
& 0.0274 & 4.3087 & 28.93 & 18.95 & 99.0 \\

+ Refine
  & 0.0253 & 3.5815 & 24.71 & 16.92 & 99.6 \\

+ Trim + Refine
  & \textbf{0.0224} & \textbf{3.1799} & \textbf{22.79} & \textbf{15.41} & \textbf{99.7} \\
  \bottomrule
  \end{tabular}%
  }
  \end{minipage}%
  \end{table}

\paragraph{Transparent-scene performance.}
Table~\ref{tab:cleargrasp_ablation} reports the results on ClearGrasp Real. Compared with the Base model, \textbf{+ Refine} improves all metrics, indicating that teacher-guided corrected supervision already helps reduce the effect of biased real sensor labels. More importantly, \textbf{+ Trim + Refine} further improves over \textbf{+ Refine}, reducing AbsRel from 0.0253 to 0.0224, RMSE from 24.71mm to 22.79mm, and MAE from 16.92mm to 15.41mm. This suggests that directly aligning teacher predictions to all valid sensor pixels can still be affected by locally biased measurements, such as background leakage, missing depth, or distorted geometry around transparent objects. By trimming high-disagreement pixels during scale-shift estimation, the teacher prediction is calibrated by more reliable real pixels, leading to cleaner corrected supervision for optically challenging regions.

\begin{table*}[h]
\caption{
Quantitative results on \textbf{generalization benchmarks} using affine-invariant depth metrics.
AbsRel and $\delta_1$ are reported in percentage.
The best values are highlighted in \textbf{bold}.
}
\label{tab:general_ablation}
\centering
\setlength{\tabcolsep}{2.4pt}
\renewcommand{\arraystretch}{1.15}
\scriptsize
\resizebox{\textwidth}{!}{%
\begin{tabular}{l | c c | c c | c c | c c | c c | c c | c c}
\toprule
\multirow{2}{*}{Method}
& \multicolumn{2}{c|}{NYUv2}
& \multicolumn{2}{c|}{ETH3D}
& \multicolumn{2}{c|}{iBims-1}
& \multicolumn{2}{c|}{Sintel}
& \multicolumn{2}{c|}{DIODE}
& \multicolumn{2}{c|}{HAMMER}
& \multicolumn{2}{c}{Mean}
\tabularnewline
& AbsRel$\downarrow$ & $\delta_1\uparrow$
& AbsRel$\downarrow$ & $\delta_1\uparrow$
& AbsRel$\downarrow$ & $\delta_1\uparrow$
& AbsRel$\downarrow$ & $\delta_1\uparrow$
& AbsRel$\downarrow$ & $\delta_1\uparrow$
& AbsRel$\downarrow$ & $\delta_1\uparrow$
& AbsRel$\downarrow$ & $\delta_1\uparrow$
\tabularnewline
\midrule
Base
& 5.230 & 96.67
& \textbf{6.917} & \textbf{94.50}
& 4.947 & 96.31
& 19.508 & 73.10
& 7.095 & 91.80
& 4.487 & 97.86
& 8.031 & 91.71
\tabularnewline

{}+ Refine
& \textbf{5.224} & \textbf{96.75}
& 6.937 & 94.23
& \textbf{4.855} & \textbf{96.58}
& 19.413 & 72.74
& \textbf{7.051} & \textbf{92.06}
& 4.229 & 98.21
& 7.951 & 91.76
\tabularnewline

{}+ Trim + Refine
& 5.273 & 96.70
& 7.025 & 94.13
& 4.930 & 96.57
& \textbf{19.368} & \textbf{73.22}
& 7.125 & 91.74
& \textbf{3.912} & \textbf{98.42}
& \textbf{7.939} & \textbf{91.80}
\tabularnewline
\bottomrule
\end{tabular}%
}
\end{table*}

\paragraph{Generalization on standard benchmarks.}
Table~\ref{tab:general_ablation} evaluates whether real-label rehabilitation affects general zero-shot depth performance. Although \textbf{+ Trim + Refine} is not the best on every individual dataset, it achieves the best mean performance across all benchmarks, with the lowest average AbsRel and the highest average $\delta_1$. This shows that residual-trimmed refinement does not simply overfit to transparent-object scenes, but improves the overall quality of real-label rehabilitation. The improvement is especially clear on HAMMER, an object-centric benchmark containing strong local structures and tabletop objects. This is consistent with our motivation: sensor-induced bias is often localized around objects, boundaries, and optically challenging regions, so trimming unreliable alignment pixels is particularly beneficial in object-centric scenarios.

\begin{table}[h]
\centering
\caption{
Boundary sharpness ablation using radius-1 F1 scores.
All values are reported in percentage.
}
\label{tab:boundary_ablation}
\tiny
\renewcommand{\arraystretch}{1.05}
\resizebox{0.5\linewidth}{!}{%
\begin{tabular}{lccc}
\toprule
\textbf{Variant}
& \textbf{iBims-1}
& \textbf{HAMMER}
& \textbf{Mean}
\tabularnewline
\midrule
Base
& 12.07
& 2.39
& 7.23
\tabularnewline

+ Refine
  & 13.77
  & 2.98
  & 8.37
  \tabularnewline

+ Trim + Refine
  & \textbf{14.33}
  & \textbf{3.93}
  & \textbf{9.13}
  \tabularnewline
  \bottomrule
  \end{tabular}%
  }
  \end{table}

\paragraph{Boundary sharpness.}
Table~\ref{tab:boundary_ablation} further evaluates local structure recovery using radius-1 boundary F1. Both \textbf{+ Refine} and \textbf{+ Trim + Refine} consistently improve F1 scores over the Base model, and \textbf{+ Trim + Refine} achieves the best results on both iBims-1 and HAMMER. This indicates that the gain is not merely caused by better global scale-shift alignment. Instead, residual-trimmed rehabilitation improves local geometric discontinuities and object boundaries, which are important for transparent objects, reflective surfaces, and embodied perception. Overall, these ablations support our claim that correcting biased real supervision with residual-trimmed teacher alignment improves optical robustness while preserving general monocular geometry capability.

\subsection{Sensitivity to Residual Trimming Ratio}

We further study the sensitivity of residual-trimmed alignment to the trimming ratio $\tau$.
All variants are trained under the same controlled setting using ARKitScenes and HyperSim, with all other training configurations kept unchanged.
We evaluate $\tau \in \{0, 0.05, 0.10, 0.15, 0.20, 0.30\}$ on ClearGrasp Real, where $\tau=0$ corresponds to teacher-guided refinement without residual trimming.
As shown in Table~\ref{tab:tau_ablation}, performance remains reasonably stable over a range of trimming ratios, with the adopted $\tau=0.10$ achieving the best overall result.
A small trimming ratio may leave more unreliable sensor measurements in the alignment set, whereas a large ratio may remove useful calibration cues.
Overall, $\tau=0.10$ provides a favorable trade-off for robust scale--shift alignment.

\begin{table}[h]
\centering
\caption{{Sensitivity to residual trimming ratio $\tau$ on ClearGrasp Real.}
All variants use the same training setup. RMSE and MAE are reported in mm.}
\label{tab:tau_ablation}

\scriptsize
\setlength{\tabcolsep}{5pt}
\renewcommand{\arraystretch}{1.05}

\begin{tabular}{ccccc}
\toprule
$\tau$
& AbsRel$\downarrow$
& SiLog$\downarrow$
& RMSE$\downarrow$
& MAE$\downarrow$ \\
\midrule
0
& 0.0253
& 3.5815
& 24.71
& 16.92 \\

0.05
& 0.0251
& 3.5248
& 24.63
& 16.89 \\

\rowcolor{gray!12}
\textbf{0.10}
& \textbf{0.0224}
& \textbf{3.1799}
& \textbf{22.79}
& \textbf{15.41} \\

0.15
& 0.0272
& 3.7743
& 27.23
& 18.68 \\

0.20
& 0.0254
& 3.4832
& 25.55
& 17.70 \\

0.30
& 0.0251
& 3.5165
& 25.14
& 17.04 \\

\bottomrule
\end{tabular}
\end{table}

\subsection{Effect of Transparency-targeted Rendering}

\begin{figure}[!t]
\centering
\includegraphics[width=0.99\textwidth]{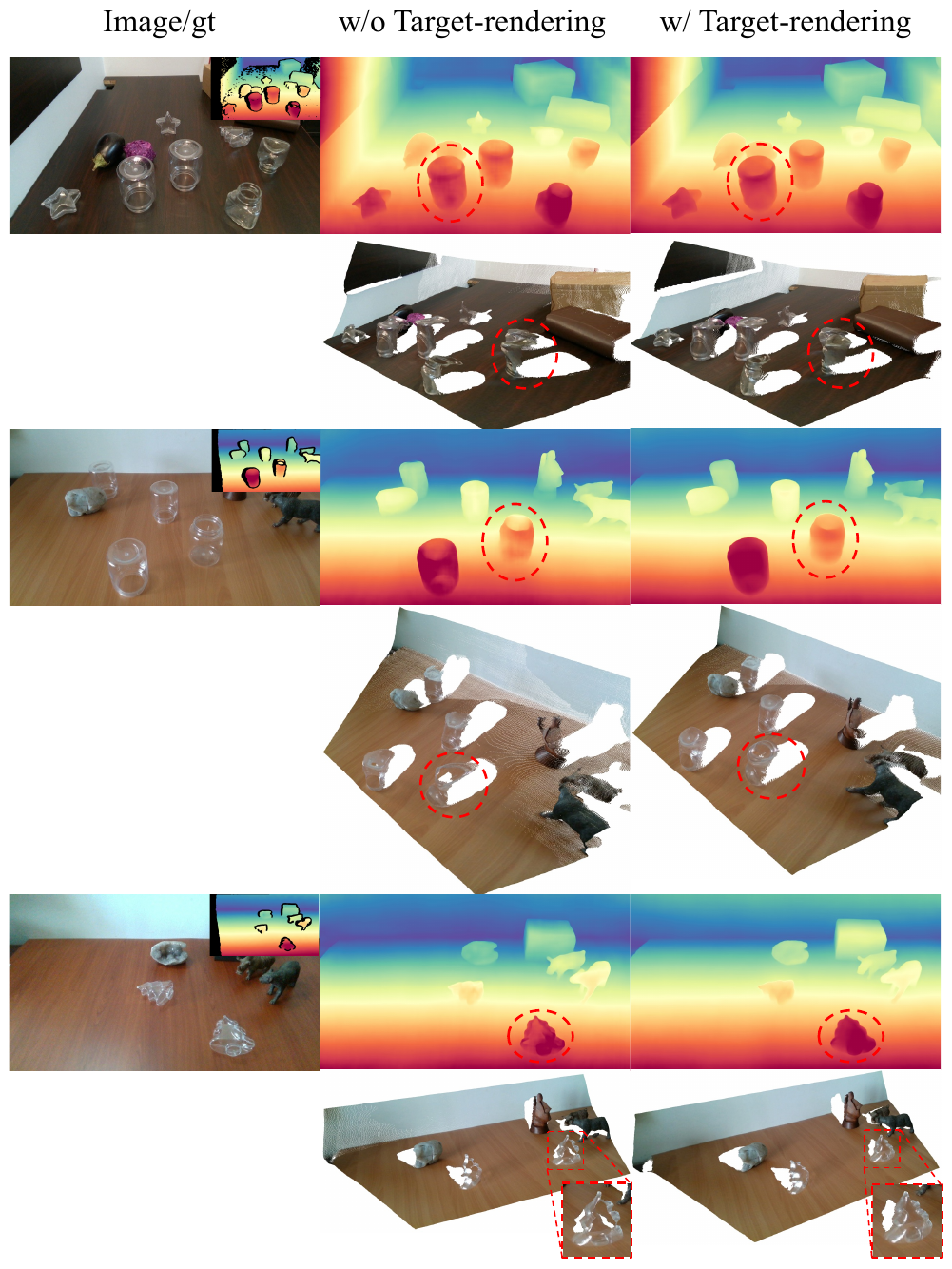}
\caption{
\textbf{Qualitative ablation of transparency-targeted rendering.}
We compare models trained with and without our targeted rendering set under the full-data setting.
Without targeted rendering, the model recovers the global tabletop layout but often misses, flattens, or distorts transparent objects.
With targeted rendering, the model produces more coherent transparent-object geometry, sharper object boundaries, and more complete point-map structures, especially in the highlighted regions.
}
\label{fig:ablation_rendering}
\end{figure}

\paragraph{Rendering ablation under full-data training.}
We further study the role of transparency-targeted rendering under the full-data training setting. In this ablation, both variants use the same training pipeline and label rehabilitation strategy, while the only difference is whether our compact targeted rendering set is included. This isolates the effect of clean optical geometry supervision from other training components.

As shown in Fig.~\ref{fig:ablation_rendering}, the model without targeted rendering can recover the coarse tabletop layout and background geometry, but transparent objects are often incomplete, over-smoothed, or absorbed into the surrounding surface. This indicates that broad real and synthetic datasets provide sufficient supervision for scene-level geometry, but still lack dense and reliable object-centric supervision for glassware, refractive boundaries, and local occlusion structures.

Adding targeted rendering substantially improves the reconstruction of transparent objects. The highlighted regions show that the model trained with targeted rendering recovers more complete cup and glass shapes, produces clearer foreground-background separation, and preserves more plausible 3D object structures in the point-map visualization. This supports our view that targeted rendering should not be treated as large-scale domain-specific fine-tuning. Instead, it serves as a compact source of clean optical geometry that complements real-label rehabilitation. Teacher-guided refinement suppresses sensor-induced bias in real supervision, while targeted rendering exposes the student to transparent-object structures that are rarely annotated reliably in real RGB-D data. These two components are therefore complementary for learning robust monocular geometry in optically challenging scenes.

\section{Efficiency Analysis}
\label{sec:efficiency}

We further evaluate the inference efficiency of OptiGeo and representative geometry baselines on ClearGrasp Real. Since the first few frames may include model initialization, CUDA warm-up, and preprocessing overhead, we report the post-cold-start average latency and FPS by excluding the initial warm-up frames. This provides a more stable estimate of practical runtime during continuous deployment.

\begin{table*}[h]
\centering
\caption{
\textbf{Efficiency comparison on ClearGrasp Real.}
We list representative runtime settings in our benchmark.
Memory denotes peak allocated GPU memory.
Latency and FPS are averaged after the initial cold-start period.
}
\label{tab:efficiency_all}
\scriptsize
\renewcommand{\arraystretch}{1.1}
\setlength{\tabcolsep}{4.0pt}
\resizebox{0.85\textwidth}{!}{%
\begin{tabular}{llccccc}
\toprule
\textbf{Method}
& \textbf{Precision}
& \textbf{Input}
& \textbf{Params (M)}
& \textbf{Mem. (GB)}
& \textbf{Latency (ms) $\downarrow$}
& \textbf{FPS $\uparrow$} \\
\midrule
OptiGeo
& fp32
& tokens3600
& \textbf{30.1}
& 1.13
& 105.50
& 9.48 \\

OptiGeo
& fp16
& tokens1000
& \textbf{30.1}
& \textbf{0.42}
& \textbf{32.77}
& \textbf{30.52} \\

\midrule
DAv3
& fp32
& native
& 350
& 2.46
& 298.99
& 3.34 \\

DAv3
& fp16
& tokens1000
& 410.9
& 2.57
& 173.78
& 5.75 \\

\midrule
VGGT
& fp16
& native
& 1200
& 6.90
& 285.83
& 3.50 \\

VGGT
& fp32
& native
& 1200
& 5.22
& 372.60
& 2.68 \\

\midrule
MoGe
& fp16
& tokens2500
& 314.2
& 2.15
& 84.16
& 11.88 \\

MoGe
& fp16
& tokens1000
& 314.2
& 2.00
& 39.18
& 25.52 \\

\midrule
UniDepthv2
& fp32
& native
& 353.8
& 2.82
& 62.33
& 16.04 \\
\bottomrule
\end{tabular}%
}
\end{table*}

Table~\ref{tab:efficiency_all} summarizes the efficiency comparison under representative runtime settings. OptiGeo achieves the best overall efficiency among all evaluated methods. Under the fp16 tokens1000 setting, OptiGeo runs at 32.77ms per frame, corresponding to 30.52 FPS, while using only 30.1M parameters and 0.42GB peak allocated GPU memory. Even under the fp32 tokens3600 setting, OptiGeo keeps the memory footprint low and remains substantially smaller than all compared baselines.

The token-controlled comparison further highlights the efficiency of the compact student architecture. Under the same fp16 tokens1000 setting, MoGe runs at 39.18ms and 25.52 FPS, while OptiGeo reaches 32.77ms and 30.52 FPS. Meanwhile, OptiGeo uses about one tenth of the parameters and less than one quarter of the GPU memory. Compared with DAv3 under the fp16 tokens1000 setting, OptiGeo is about $5.3\times$ faster and uses much less memory. These results indicate that the runtime gain of OptiGeo does not only come from token control or input preprocessing, but also from its lightweight model design.

Compared with large geometry foundation models, the efficiency gap is more significant. VGGT requires more than one billion parameters and substantially higher memory, while its fp16 inference remains much slower than OptiGeo. UniDepthv2 is the fastest non-OptiGeo baseline in our benchmark, but still has nearly twice the latency of OptiGeo and requires a much larger model. This demonstrates that scaling model size can improve general geometry representation, but it also introduces a considerable deployment cost.

This efficiency advantage is important for embodied perception, where monocular geometry estimation usually runs together with detection, mapping, planning, or control modules under limited memory and latency budgets. In OptiGeo, the clean-geometry teacher and label rehabilitation pipeline are used only during training. At inference time, the system deploys only the compact 30M feed-forward student. Therefore, OptiGeo improves optical robustness without adding inference-time modules, making it suitable for real-time robotic perception in transparent and reflective scenes.

\section{Detailed Loss Design}
\label{sec:loss}

We adopt a set of affine-invariant and geometry-aware losses to train a strong relative geometry base model.
The network predicts a relative point map $\hat{\mathbf P}\in\mathbb{R}^{H\times W\times 3}$ and a reliability mask $\hat{\mathbf M}\in[0,1]^{H\times W}$, and is optimized with complementary global, local, normal, edge, and mask supervision.
The target point map is denoted as $\mathbf P\in\mathbb{R}^{H\times W\times 3}$.
For the $i$-th pixel, we denote the predicted and target 3D points as $\hat{\mathbf p}_i$ and $\mathbf p_i$, respectively.
The valid supervision set is denoted as $\mathcal M$.

\vspace{0.05in}\noindent \textbf{Global Alignment Loss.}
The global alignment loss fits $\hat{\mathbf P}$ to the target point map under a global affine ambiguity.
Specifically, we estimate a global scale and translation $(s^*,\mathbf t^*)$ that align the predicted relative point map to the target point map.
The global affine-invariant loss is defined as
\begin{equation}
\mathcal L_{\rm global}
=
\sum_{i\in \mathcal M}
\frac{1}{z_i}
\left\|
s^*\hat{\mathbf p}_i+\mathbf t^*-\mathbf p_i
\right\|_1,
\label{eq:global_loss_sup}
\end{equation}
where $z_i$ is the $z$-coordinate of $\mathbf p_i$.
In practice, $(s^*,\mathbf t^*)$ are estimated by solving a least-squares alignment between the prediction and the target point map over valid pixels.
The weighting term $1/z_i$ balances the supervision strength across large depth ranges and prevents far-range points from dominating the objective.

This objective encourages the model to recover globally consistent relative geometry while remaining agnostic to absolute metric scale.
When combined with the local losses below, it provides a coarse-to-fine training signal for point-map learning.

\vspace{0.05in}\noindent \textbf{Multi-scale Local Patch Losses.}
To preserve local structures and high-frequency details, we additionally apply multi-scale local patch supervision.
The overall local objective is
\begin{equation}
\mathcal L_{\rm local}
=
\sum_{\alpha \in \mathcal A}
\mathcal L_{S(\alpha)},
\label{eq:local_loss_total_sup}
\end{equation}
where $\mathcal A$ is the set of local neighborhood scales.

For each scale $\alpha$, we construct a local spherical neighborhood around an anchor point $\mathbf p_j$:
\begin{equation}
\mathcal S_j
=
\left\{
i \;\middle|\;
\|\mathbf p_i-\mathbf p_j\|\le r_j,\; i\in\mathcal M
\right\},
\label{eq:sphere_region_sup}
\end{equation}
with radius
\begin{equation}
r_j
=
\alpha \cdot z_j \cdot \frac{\sqrt{W^2+H^2}}{2f},
\label{eq:radius_sup}
\end{equation}
where $z_j$ is the depth of $\mathbf p_j$, $f$ is the ground-truth focal length, and $(W,H)$ is the image resolution.
This design makes the neighborhood size adaptive to both scene depth and camera intrinsics.

A local affine alignment is then solved within each spherical patch using scale and translation parameters $(s_j^*,\mathbf t_j^*)$.
The patch loss is defined as
\begin{equation}
\mathcal L_{S(\alpha)}
=
\sum_{j\in\mathcal H_\alpha}
\sum_{i\in\mathcal S_j}
\frac{1}{z_i}
\left\|
s_j^* \hat{\mathbf p}_i + \mathbf t_j^* - \mathbf p_i
\right\|_1,
\label{eq:local_loss_sup}
\end{equation}
where $\mathcal H*\alpha$ denotes the set of anchors sampled at scale $\alpha$.
Compared with the global loss, the local patch losses allow independent affine alignment in local regions, which better preserves object boundaries, thin structures, and local depth discontinuities.

In practice, we use different local scales for different supervision sources.
Synthetic data uses three levels $\mathcal A_{\rm syn}={\tfrac{1}{4}, \tfrac{1}{16}, \tfrac{1}{64}}$, SfM data uses $\mathcal A_{\rm sfm}={\tfrac{1}{4}, \tfrac{1}{16}}$, and LiDAR data uses only $\mathcal A_{\rm lidar}={\tfrac{1}{4}}$.

\vspace{0.05in}\noindent \textbf{Surface-Normal Consistency Loss.}
To encourage piecewise smooth yet detail-preserving geometry, we impose a surface-normal consistency loss:
\begin{equation}
\mathcal L_{\rm normal}
=
\sum_{i\in\mathcal M}
\angle(\hat{\mathbf n}_i,\mathbf n_i),
\label{eq:normal_loss_sup}
\end{equation}
where $\hat{\mathbf n}_i$ and $\mathbf n_i$ denote the predicted and target surface normals, respectively, and $\angle(\cdot,\cdot)$ measures their angular discrepancy.
This loss encourages locally coherent surface orientation and complements the point-wise alignment losses.

\noindent \textbf{Edge Loss.}
To better preserve geometric discontinuities and local structure, we further supervise the directions of neighboring 3D point differences. Let
\begin{equation}
\Delta_x \hat{\mathbf p}_{u,v} = \hat{\mathbf p}_{u,v} - \hat{\mathbf p}_{u+1,v},
\qquad
\Delta_y \hat{\mathbf p}_{u,v} = \hat{\mathbf p}_{u,v} - \hat{\mathbf p}_{u,v+1},
\end{equation}
denote the horizontal and vertical edge vectors of the predicted point map, and define $\Delta_x \mathbf p_{u,v}$ and $\Delta_y \mathbf p_{u,v}$ similarly for the ground truth. We then penalize their angular discrepancy:
\begin{equation}
\mathcal L_{\rm edge}
=
\sum_{(u,v)\in\mathcal M_x}
\angle(\Delta_x \hat{\mathbf p}_{u,v}, \Delta_x \mathbf p_{u,v})
+
\sum_{(u,v)\in\mathcal M_y}
\angle(\Delta_y \hat{\mathbf p}_{u,v}, \Delta_y \mathbf p_{u,v}),
\label{eq:edge_loss_sup}
\end{equation}
where $\mathcal M_x$ and $\mathcal M_y$ denote the valid neighboring pixel pairs in the horizontal and vertical directions, respectively. This loss encourages the predicted point map to preserve local edge directions and sharp geometric transitions.

\vspace{0.05in}\noindent \textbf{Mask Supervision Loss.}
To suppress unreliable regions, we supervise the predicted reliability mask with
\begin{equation}
\mathcal L_{\rm mask}
=
\left\|
\hat{\mathbf M} - \left(1-\mathbf M_{\rm inf}\right)
\right\|_2^2,
\label{eq:mask_loss_sup}
\end{equation}
where $\mathbf M*{\rm inf}$ denotes the invalid-region indicator derived from the target annotation.
The mask loss encourages the model to identify valid geometric regions and avoid over-confident predictions on invalid or unreliable supervision.

\vspace{0.05in}\noindent \textbf{Overall Objective.}
The final loss is label-type dependent:
\begin{equation}
\mathcal L_{\rm relative}^{(t)}
=
\lambda_{\rm g}^{(t)}\mathcal L_{\rm global}
+
\sum_{\alpha\in\mathcal A_t}
\lambda_{\alpha}^{(t)}\mathcal L_{S(\alpha)}
+
\lambda_{\rm n}^{(t)}\mathcal L_{\rm normal}
+
\lambda_{\rm e}^{(t)}\mathcal L_{\rm edge}
+
\lambda_{\rm m}^{(t)}\mathcal L_{\rm mask},
\label{eq:loss_relative_sup}
\end{equation}
where $t\in{\texttt{synthetic},\texttt{sfm},\texttt{lidar}}$ denotes the label type.

% In all experiments, we set the loss weights as follows:
% \begin{equation}
% \begin{aligned}
% &(\lambda_{\rm g}^{(t)}, \lambda_{1/4}^{(t)}, \lambda_{1/16}^{(t)}, 
% \lambda_{1/64}^{(t)}, \lambda_{\rm n}^{(t)}, \lambda_{\rm e}^{(t)}, 
% \lambda_{\rm m}^{(t)}) \\
% &\quad =
% \begin{cases}
% (1, 1, 1, 1, 0.1, 1, 0.1), & t=\mathrm{synthetic},\\
% (1, 1, 1, 0, 0.1, 1, 0.1), & t=\mathrm{sfm},\\
% (1, 1, 0, 0, 0, 0, 0), & t=\mathrm{lidar}.
% \end{cases}
% \end{aligned}
% \label{eq:relative_weights_sup}
% \end{equation}

Unless otherwise stated, the global and local alignment terms all use unit weight.
In implementation, the global loss uses an alignment resolution of $48$, while the local patch losses at levels ${4,16,64}$ use alignment resolutions ${24,12,6}$ and numbers of sampled patches ${16,256,4096}$, respectively.

\section{Targeted Data Generation Details}
\label{sec:data_generation}

We generate the transparency-targeted rendering set using Infinigen. The rendered data contains two complementary subsets: an object-centric optical subset and a general-scene subset. The object-centric subset focuses on transparent tabletop objects and provides clean local geometry supervision for glassware, refractive boundaries, and object-level occlusion structures. The general-scene subset preserves broader indoor geometric diversity and prevents the model from over-specializing to tabletop transparent-object layouts.

\begin{table}[h]
\centering
\caption{
\textbf{Details of the transparency-targeted rendering set.}
The object-centric subset provides clean supervision for transparent glassware geometry, while the general-scene subset maintains broader scene-level diversity.
}
\label{tab:rendering_details}
\scriptsize
\renewcommand{\arraystretch}{1.15}
\setlength{\tabcolsep}{4.0pt}
\resizebox{0.88\linewidth}{!}{%
\begin{tabular}{lcccl}
\toprule
\textbf{Subset}
& \textbf{\# Scenes}
& \textbf{\# Images}
& \textbf{Main Content}
& \textbf{Purpose} \\

\midrule
Object-centric optical
& 571
& 4,564
& Tabletop glassware
& Clean transparent-object geometry \\

General-scene
& 395
& 3,106
& Indoor layouts
& Scene-level geometric diversity \\
\midrule

Total
& 966
& 7,670
& --
& Targeted optical geometry supervision \\
\bottomrule
\end{tabular}%
}
\end{table}

For the object-centric optical subset, each scene is initialized as a single-room indoor dining-room environment with terrain generation disabled. We insert a randomized set of transparent glassware objects, including 3--6 cups and 3--6 wineglasses. The minimum inter-object spacing is set to 0.12m to reduce severe overlap while still preserving object interactions and occlusions. All target objects are forced to use glass material, and an additional focus light is added to improve the visibility of transparent boundaries.

For each object-centric scene, we sample orbiting camera views around the tabletop. The camera distance is randomly sampled from 1.2--2.0m, the camera height from 0.3--0.6m, and the orbit start angle from 0--45 degrees. Images are rendered at $1024 \times 768$ resolution using a $24\mathrm{mm} \times 18\mathrm{mm}$ sensor model. The focal length is randomly sampled from 23.44--32.11mm, corresponding to approximately 1000--1370 pixels. RGB images are rendered with Cycles using up to 1024 samples, while annotation maps are rendered using a flat rendering pass. After filtering, this subset contains 4,564 images from 571 scenes.

The general-scene subset is generated to complement the object-centric data. It contains 3,106 images from 395 indoor scenes, providing more diverse room layouts, object arrangements, depth ranges, and background structures. This subset is important because optical robustness should be learned as part of a general monocular geometry model, rather than as a narrow transparent-tabletop specialization. Together, the two subsets provide compact but targeted clean supervision: the object-centric subset teaches local transparent-object geometry, while the general-scene subset regularizes the model with broader scene-level geometry.

\section{Teacher Implementation Details}
\label{sec}
\paragraph{Teacher architecture.}
The clean-geometry teacher follows the same overall formulation as the student model. It predicts a dense relative point map and uses the same decoder design as the student, but adopts a larger DINOv3 Hplus initialized encoder for stronger geometric representation. The teacher has approximately 940M parameters. This design keeps the teacher and student geometrically consistent: the teacher provides corrected point-map/depth supervision during training, while the compact student learns the same prediction target for efficient inference.
\paragraph{Teacher training.}
To avoid inheriting real sensor bias, the teacher is trained only on synthetic data with reliable geometric supervision. We train the teacher for 50K iterations using 32 NVIDIA H20 GPUs. Since the teacher is used only to generate corrected supervision for real images, it is not deployed during inference. This separation allows OptiGeo to benefit from a strong clean-geometry prior during training while preserving a compact 30M-parameter student at test time.
\paragraph{Real-label rehabilitation.}
After training the synthetic-only teacher, we use it to rehabilitate real sensor labels. We apply teacher-guided refinement to all real training datasets except STD and BlendedMVS. Given a real image and its raw sensor depth, the frozen teacher first predicts a relative depth map from the z-axis of its point map. We then align the teacher prediction to the raw sensor depth using the residual-trimmed scale-shift fitting described in the main paper.
\paragraph{Residual-trimmed alignment.}
The trimming ratio is set to $\tau=10\%$ in all experiments. For each real image, we first estimate an initial global scale and shift between the teacher depth and the raw sensor depth over valid sensor pixels. We then compute the absolute residuals and remove the top $10\%$ high-residual pixels when re-estimating the final scale and shift. These high-residual pixels often correspond to unreliable sensor measurements, such as background leakage on glass, missing transparent-object depth, or distorted reflective regions. Importantly, the trimming operation is used only for estimating the alignment parameters. After the final scale and shift are obtained, the aligned teacher prediction is used as dense corrected supervision.
\paragraph{Student supervision.}
For rehabilitated real samples, the raw biased depth label is replaced by the aligned teacher depth. The corrected depth is then converted into point-map supervision and used in the same training objective as other samples. In this way, real data still contributes realistic appearance, scene layout, and metric anchoring, while locally unreliable sensor labels are suppressed by the clean-geometry teacher. The final deployed model is only the student; the teacher and rehabilitation pipeline introduce no inference-time cost.

%% file: example.bib
@inproceedings{baruch1arkitscenes,
  title={ARKitScenes: A Diverse Real-World Dataset For 3D Indoor Scene Understanding Using Mobile RGB-D Data},
  author={Baruch, Gilad and Chen, Zhuoyuan and Dehghan, Afshin and Feigin, Yuri and Fu, Peter and Gebauer, Thomas and Kurz, Daniel and Dimry, Tal and Joffe, Brandon and Schwartz, Arik and others},
  booktitle={Thirty-fifth Conference on Neural Information Processing Systems Datasets and Benchmarks Track (Round 1)}
}

@inproceedings{yao2020blendedmvs,
  title={Blendedmvs: A large-scale dataset for generalized multi-view stereo networks},
  author={Yao, Yao and Luo, Zixin and Li, Shiwei and Zhang, Jingyang and Ren, Yufan and Zhou, Lei and Fang, Tian and Quan, Long},
  booktitle={Proceedings of the IEEE/CVF conference on computer vision and pattern recognition},
  pages={1790--1799},
  year={2020}
}

@inproceedings{zamir2018taskonomy,
  title={Taskonomy: Disentangling task transfer learning},
  author={Zamir, Amir R and Sax, Alexander and Shen, William and Guibas, Leonidas J and Malik, Jitendra and Savarese, Silvio},
  booktitle={Proceedings of the IEEE conference on computer vision and pattern recognition},
  pages={3712--3722},
  year={2018}
}

@inproceedings{sun2020scalability,
  title={Scalability in perception for autonomous driving: Waymo open dataset},
  author={Sun, Pei and Kretzschmar, Henrik and Dotiwalla, Xerxes and Chouard, Aurelien and Patnaik, Vijaysai and Tsui, Paul and Guo, James and Zhou, Yin and Chai, Yuning and Caine, Benjamin and others},
  booktitle={Proceedings of the IEEE/CVF conference on computer vision and pattern recognition},
  pages={2446--2454},
  year={2020}
}

@inproceedings{roberts2021hypersim,
  title={Hypersim: A photorealistic synthetic dataset for holistic indoor scene understanding},
  author={Roberts, Mike and Ramapuram, Jason and Ranjan, Anurag and Kumar, Atulit and Bautista, Miguel Angel and Paczan, Nathan and Webb, Russ and Susskind, Joshua M},
  booktitle={Proceedings of the IEEE/CVF international conference on computer vision},
  pages={10912--10922},
  year={2021}
}

@article{wang2019irs,
  title={Irs: A large naturalistic indoor robotics stereo dataset to train deep models for disparity and surface normal estimation},
  author={Wang, Qiang and Zheng, Shizhen and Yan, Qingsong and Deng, Fei and Zhao, Kaiyong and Chu, Xiaowen},
  journal={arXiv preprint arXiv:1912.09678},
  year={2019}
}

@article{niklaus20193d,
  title={3d ken burns effect from a single image},
  author={Niklaus, Simon and Mai, Long and Yang, Jimei and Liu, Feng},
  journal={ACM Transactions on Graphics (ToG)},
  volume={38},
  number={6},
  pages={1--15},
  year={2019},
  publisher={ACM New York, NY, USA}
}

@inproceedings{li2023matrixcity,
  title={Matrixcity: A large-scale city dataset for city-scale neural rendering and beyond},
  author={Li, Yixuan and Jiang, Lihan and Xu, Linning and Xiangli, Yuanbo and Wang, Zhenzhi and Lin, Dahua and Dai, Bo},
  booktitle={Proceedings of the IEEE/CVF International Conference on Computer Vision},
  pages={3205--3215},
  year={2023}
}

@inproceedings{fonder2019mid,
  title={Mid-air: A multi-modal dataset for extremely low altitude drone flights},
  author={Fonder, Michael and Van Droogenbroeck, Marc},
  booktitle={Proceedings of the IEEE/CVF conference on computer vision and pattern recognition workshops},
  pages={0--0},
  year={2019}
}

@inproceedings{huang2018deepmvs,
  title={Deepmvs: Learning multi-view stereopsis},
  author={Huang, Po-Han and Matzen, Kevin and Kopf, Johannes and Ahuja, Narendra and Huang, Jia-Bin},
  booktitle={Proceedings of the IEEE conference on computer vision and pattern recognition},
  pages={2821--2830},
  year={2018}
}

@inproceedings{mehl2023spring,
  title={Spring: A high-resolution high-detail dataset and benchmark for scene flow, optical flow and stereo},
  author={Mehl, Lukas and Schmalfuss, Jenny and Jahedi, Azin and Nalivayko, Yaroslava and Bruhn, Andr{\'e}s},
  booktitle={Proceedings of the IEEE/CVF Conference on Computer Vision and Pattern Recognition},
  pages={4981--4991},
  year={2023}
}

@inproceedings{zheng2020structured3d,
  title={Structured3d: A large photo-realistic dataset for structured 3d modeling},
  author={Zheng, Jia and Zhang, Junfei and Li, Jing and Tang, Rui and Gao, Shenghua and Zhou, Zihan},
  booktitle={European Conference on Computer Vision},
  pages={519--535},
  year={2020},
  organization={Springer}
}

@inproceedings{wang2020tartanair,
  title={Tartanair: A dataset to push the limits of visual slam},
  author={Wang, Wenshan and Zhu, Delong and Wang, Xiangwei and Hu, Yaoyu and Qiu, Yuheng and Wang, Chen and Hu, Yafei and Kapoor, Ashish and Scherer, Sebastian},
  booktitle={2020 IEEE/RSJ International Conference on Intelligent Robots and Systems (IROS)},
  pages={4909--4916},
  year={2020},
  organization={IEEE}
}

@article{gomez2025all,
  title={All for one, and one for all: Urbansyn dataset, the third musketeer of synthetic driving scenes},
  author={G{\'o}mez, Jose L and Silva, Manuel and Seoane, Antonio and Borr{\'a}s, Agn{\`e}s and Noriega, Mario and Ros, Germ{\'a}n and Iglesias-Guitian, Jose A and L{\'o}pez, Antonio M},
  journal={Neurocomputing},
  volume={637},
  pages={130038},
  year={2025},
  publisher={Elsevier}
}

@article{karaev2023dynamicstereo,
  title={DynamicStereo: Consistent Dynamic Depth from Stereo Videos},
  author={Nikita Karaev and Ignacio Rocco and Benjamin Graham and Natalia Neverova and Andrea Vedaldi and Christian Rupprecht},
  journal={CVPR},
  year={2023}
}

@inproceedings{dosovitskiy2021imageworth16x16words,
  title={An Image is Worth 16x16 Words: Transformers for Image Recognition at Scale},
  author={Dosovitskiy, Alexey and Beyer, Lucas and Kolesnikov, Alexander and Weissenborn, Dirk and Zhai, Xiaohua and Unterthiner, Thomas and Dehghani, Mostafa and Minderer, Matthias and Heigold, G and Gelly, S and others},
  booktitle={International Conference on Learning Representations},
  year={2020}
}

@article{simeoni2025dinov3,
  title={Dinov3},
  author={Sim{\'e}oni, Oriane and Vo, Huy V and Seitzer, Maximilian and Baldassarre, Federico and Oquab, Maxime and Jose, Cijo and Khalidov, Vasil and Szafraniec, Marc and Yi, Seungeun and Ramamonjisoa, Micha{\"e}l and others},
  journal={arXiv preprint arXiv:2508.10104},
  year={2025}
}

@inproceedings{wang2025moge,
  title={Moge: Unlocking accurate monocular geometry estimation for open-domain images with optimal training supervision},
  author={Wang, Ruicheng and Xu, Sicheng and Dai, Cassie and Xiang, Jianfeng and Deng, Yu and Tong, Xin and Yang, Jiaolong},
  booktitle={Proceedings of the Computer Vision and Pattern Recognition Conference},
  pages={5261--5271},
  year={2025}
}

@inproceedings{piccinelli2024unidepth,
  title={UniDepth: Universal monocular metric depth estimation},
  author={Piccinelli, Luigi and Yang, Yung-Hsu and Sakaridis, Christos and Segu, Mattia and Li, Siyuan and Van Gool, Luc and Yu, Fisher},
  booktitle={Proceedings of the IEEE/CVF Conference on Computer Vision and Pattern Recognition},
  pages={10106--10116},
  year={2024}
}

@article{piccinelli2025unidepthv2,
  title={Unidepthv2: Universal monocular metric depth estimation made simpler},
  author={Piccinelli, Luigi and Sakaridis, Christos and Yang, Yung-Hsu and Segu, Mattia and Li, Siyuan and Abbeloos, Wim and Van Gool, Luc},
  journal={arXiv preprint arXiv:2502.20110},
  year={2025}
}

@inproceedings{bochkovskiydepth,
  title={Depth Pro: Sharp Monocular Metric Depth in Less Than a Second},
  author={Bochkovskiy, Alexey and Delaunoy, Ama{\"e}l and Germain, Hugo and Santos, Marcel and Zhou, Yichao and Richter, Stephan and Koltun, Vladlen},
  booktitle={The Thirteenth International Conference on Learning Representations}
}

@inproceedings{wang2025moge2,
  title={MoGe-2: Accurate Monocular Geometry with Metric Scale and Sharp Details},
  author={Wang, Ruicheng and Xu, Sicheng and Dong, Yue and Deng, Yu and Xiang, Jianfeng and Lv, Zelong and Sun, Guangzhong and Tong, Xin and Yang, Jiaolong},
  booktitle={The Thirty-ninth Annual Conference on Neural Information Processing Systems}
}

@article{bhat2023zoedepth,
  title={Zoedepth: Zero-shot transfer by combining relative and metric depth},
  author={Bhat, Shariq Farooq and Birkl, Reiner and Wofk, Diana and Wonka, Peter and M{\"u}ller, Matthias},
  journal={arXiv preprint arXiv:2302.12288},
  year={2023}
}

@inproceedings{yang2024depth,
  title={Depth anything: Unleashing the power of large-scale unlabeled data},
  author={Yang, Lihe and Kang, Bingyi and Huang, Zilong and Xu, Xiaogang and Feng, Jiashi and Zhao, Hengshuang},
  booktitle={Proceedings of the IEEE/CVF conference on computer vision and pattern recognition},
  pages={10371--10381},
  year={2024}
}

@article{yang2024depth2,
  title={Depth anything v2},
  author={Yang, Lihe and Kang, Bingyi and Huang, Zilong and Zhao, Zhen and Xu, Xiaogang and Feng, Jiashi and Zhao, Hengshuang},
  journal={Advances in Neural Information Processing Systems},
  volume={37},
  pages={21875--21911},
  year={2024}
}

@article{hu2024metric3d,
  title={Metric3d v2: A versatile monocular geometric foundation model for zero-shot metric depth and surface normal estimation},
  author={Hu, Mu and Yin, Wei and Zhang, Chi and Cai, Zhipeng and Long, Xiaoxiao and Chen, Hao and Wang, Kaixuan and Yu, Gang and Shen, Chunhua and Shen, Shaojie},
  journal={IEEE Transactions on Pattern Analysis and Machine Intelligence},
  year={2024},
  publisher={IEEE}
}

@inproceedings{yin2023metric3d,
  title={Metric3d: Towards zero-shot metric 3d prediction from a single image},
  author={Yin, Wei and Zhang, Chi and Chen, Hao and Cai, Zhipeng and Yu, Gang and Wang, Kaixuan and Chen, Xiaozhi and Shen, Chunhua},
  booktitle={Proceedings of the IEEE/CVF international conference on computer vision},
  pages={9043--9053},
  year={2023}
}

@inproceedings{wang2025vggt,
  title={Vggt: Visual geometry grounded transformer},
  author={Wang, Jianyuan and Chen, Minghao and Karaev, Nikita and Vedaldi, Andrea and Rupprecht, Christian and Novotny, David},
  booktitle={Proceedings of the Computer Vision and Pattern Recognition Conference},
  pages={5294--5306},
  year={2025}
}

@inproceedings{silberman2012indoor,
  title={Indoor segmentation and support inference from rgbd images},
  author={Silberman, Nathan and Hoiem, Derek and Kohli, Pushmeet and Fergus, Rob},
  booktitle={European conference on computer vision},
  pages={746--760},
  year={2012},
  organization={Springer}
}

@inproceedings{schops2019bad,
  title={Bad slam: Bundle adjusted direct rgb-d slam},
  author={Schops, Thomas and Sattler, Torsten and Pollefeys, Marc},
  booktitle={Proceedings of the IEEE/CVF Conference on Computer Vision and Pattern Recognition},
  pages={134--144},
  year={2019}
}

@inproceedings{koch2018evaluation,
  title={Evaluation of cnn-based single-image depth estimation methods},
  author={Koch, Tobias and Liebel, Lukas and Fraundorfer, Friedrich and Korner, Marco},
  booktitle={Proceedings of the European Conference on Computer Vision (ECCV) Workshops},
  pages={0--0},
  year={2018}
}

@article{koch2020comparison,
  title={Comparison of monocular depth estimation methods using geometrically relevant metrics on the IBims-1 dataset},
  author={Koch, Tobias and Liebel, Lukas and K{\"o}rner, Marco and Fraundorfer, Friedrich},
  journal={Computer Vision and Image Understanding},
  volume={191},
  pages={102877},
  year={2020},
  publisher={Elsevier}
}

@inproceedings{butler2012naturalistic,
  title={A naturalistic open source movie for optical flow evaluation},
  author={Butler, Daniel J and Wulff, Jonas and Stanley, Garrett B and Black, Michael J},
  booktitle={European conference on computer vision},
  pages={611--625},
  year={2012},
  organization={Springer}
}

@article{vasiljevic2019diode,
  title={Diode: A dense indoor and outdoor depth dataset},
  author={Vasiljevic, Igor and Kolkin, Nick and Zhang, Shanyi and Luo, Ruotian and Wang, Haochen and Dai, Falcon Z and Daniele, Andrea F and Mostajabi, Mohammadreza and Basart, Steven and Walter, Matthew R and others},
  journal={arXiv preprint arXiv:1908.00463},
  year={2019}
}

@inproceedings{jung2023importance,
  title={On the importance of accurate geometry data for dense 3d vision tasks},
  author={Jung, HyunJun and Ruhkamp, Patrick and Zhai, Guangyao and Brasch, Nikolas and Li, Yitong and Verdie, Yannick and Song, Jifei and Zhou, Yiren and Armagan, Anil and Ilic, Slobodan and others},
  booktitle={Proceedings of the IEEE/CVF Conference on Computer Vision and Pattern Recognition},
  pages={780--791},
  year={2023}
}

@article{oquab2024dinov2,
  title={DINOv2: Learning Robust Visual Features without Supervision},
  author={Oquab, Maxime and Darcet, Timoth{\'e}e and Moutakanni, Th{\'e}o and Vo, Huy and Szafraniec, Marc and Khalidov, Vasil and Fernandez, Pierre and Haziza, Daniel and Massa, Francisco and El-Nouby, Alaaeldin and others},
  journal={Transactions on Machine Learning Research Journal},
  pages={1--31},
  year={2024}
}

@inproceedings{murai2025mast3r,
  title={MASt3R-SLAM: Real-time dense SLAM with 3D reconstruction priors},
  author={Murai, Riku and Dexheimer, Eric and Davison, Andrew J},
  booktitle={Proceedings of the Computer Vision and Pattern Recognition Conference},
  pages={16695--16705},
  year={2025}
}

@inproceedings{ranftl2021vision,
  title={Vision transformers for dense prediction},
  author={Ranftl, Ren{\'e} and Bochkovskiy, Alexey and Koltun, Vladlen},
  booktitle={Proceedings of the IEEE/CVF international conference on computer vision},
  pages={12179--12188},
  year={2021}
}

@article{ke2025marigold,
  title={Marigold: Affordable Adaptation of Diffusion-Based Image Generators for Image Analysis},
  author={Ke, Bingxin and Qu, Kevin and Wang, Tianfu and Metzger, Nando and Huang, Shengyu and Li, Bo and Obukhov, Anton and Schindler, Konrad},
  journal={arXiv preprint arXiv:2505.09358},
  year={2025}
}

@article{wen2025stereo,
  title={FoundationStereo: Zero-Shot Stereo Matching},
  author={Bowen Wen and Matthew Trepte and Joseph Aribido and Jan Kautz and Orazio Gallo and Stan Birchfield},
  journal={CVPR},
  year={2025}
}

@article{lin2025depth,
  title={Depth anything 3: Recovering the visual space from any views},
  author={Lin, Haotong and Chen, Sili and Liew, Junhao and Chen, Donny Y and Li, Zhenyu and Shi, Guang and Feng, Jiashi and Kang, Bingyi},
  journal={arXiv preprint arXiv:2511.10647},
  year={2025}
}

@inproceedings{xupixel,
  title={Pixel-Perfect Depth with Semantics-Prompted Diffusion Transformers},
  author={Xu, Gangwei and Lin, Haotong and Luo, Hongcheng and Wang, Xianqi and Yao, Jingfeng and Zhu, Lianghui and Pu, Yuechuan and Chi\_, Cheng and Sun, Haiyang and WANG, BING and others},
  booktitle={The Thirty-ninth Annual Conference on Neural Information Processing Systems}
}

@article{xu2025diffusion,
  title={Diffusion Knows Transparency: Repurposing Video Diffusion for Transparent Object Depth and Normal Estimation},
  author={Xu, Shaocong and Wei, Songlin and Wei, Qizhe and Geng, Zheng and Li, Hong and Shen, Licheng and Sun, Qianpu and Han, Shu and Ma, Bin and Li, Bohan and others},
  journal={arXiv preprint arXiv:2512.23705},
  year={2025}
}

@inproceedings{fu2024geowizard,
  title={Geowizard: Unleashing the diffusion priors for 3d geometry estimation from a single image},
  author={Fu, Xiao and Yin, Wei and Hu, Mu and Wang, Kaixuan and Ma, Yuexin and Tan, Ping and Shen, Shaojie and Lin, Dahua and Long, Xiaoxiao},
  booktitle={European Conference on Computer Vision},
  pages={241--258},
  year={2024},
  organization={Springer}
}

@inproceedings{wang2024digging,
  title={Digging into contrastive learning for robust depth estimation with diffusion models},
  author={Wang, Jiyuan and Lin, Chunyu and Nie, Lang and Liao, Kang and Shao, Shuwei and Zhao, Yao},
  booktitle={Proceedings of the 32nd ACM International Conference on Multimedia},
  pages={4129--4137},
  year={2024}
}

@inproceedings{liu2025monocular,
  title={Monocular depth estimation and segmentation for transparent object with iterative semantic and geometric fusion},
  author={Liu, Jiangyuan and Ma, Hongxuan and Guo, Yuxin and Zhao, Yuhao and Zhang, Chi and Sui, Wei and Zou, Wei},
  booktitle={2025 IEEE International Conference on Robotics and Automation (ICRA)},
  pages={11162--11168},
  year={2025},
  organization={IEEE}
}

@inproceedings{costanzino2023learning,
  title={Learning depth estimation for transparent and mirror surfaces},
  author={Costanzino, Alex and Ramirez, Pierluigi Zama and Poggi, Matteo and Tosi, Fabio and Mattoccia, Stefano and Di Stefano, Luigi},
  booktitle={Proceedings of the IEEE/CVF International Conference on Computer Vision},
  pages={9244--9255},
  year={2023}
}

@inproceedings{yeshwanth2023scannet++,
  title={Scannet++: A high-fidelity dataset of 3d indoor scenes},
  author={Yeshwanth, Chandan and Liu, Yueh-Cheng and Nie{\ss}ner, Matthias and Dai, Angela},
  booktitle={Proceedings of the IEEE/CVF International Conference on Computer Vision},
  pages={12--22},
  year={2023}
}

@inproceedings{dai2022domain,
  title={Domain randomization-enhanced depth simulation and restoration for perceiving and grasping specular and transparent objects},
  author={Dai, Qiyu and Zhang, Jiyao and Li, Qiwei and Wu, Tianhao and Dong, Hao and Liu, Ziyuan and Tan, Ping and Wang, He},
  booktitle={European Conference on Computer Vision},
  pages={374--391},
  year={2022},
  organization={Springer}
}

@inproceedings{zheng2023pointodyssey,
  title={Pointodyssey: A large-scale synthetic dataset for long-term point tracking},
  author={Zheng, Yang and Harley, Adam W and Shen, Bokui and Wetzstein, Gordon and Guibas, Leonidas J},
  booktitle={Proceedings of the IEEE/CVF International Conference on Computer Vision},
  pages={19855--19865},
  year={2023}
}

@inproceedings{sajjan2020clear,
  title={Clear grasp: 3d shape estimation of transparent objects for manipulation},
  author={Sajjan, Shreeyak and Moore, Matthew and Pan, Mike and Nagaraja, Ganesh and Lee, Johnny and Zeng, Andy and Song, Shuran},
  booktitle={2020 IEEE international conference on robotics and automation (ICRA)},
  pages={3634--3642},
  year={2020},
  organization={IEEE}
}

@article{fang2022transcg,
  title={Transcg: A large-scale real-world dataset for transparent object depth completion and a grasping baseline},
  author={Fang, Hongjie and Fang, Hao-Shu and Xu, Sheng and Lu, Cewu},
  journal={IEEE Robotics and Automation Letters},
  volume={7},
  number={3},
  pages={7383--7390},
  year={2022},
  publisher={IEEE}
}

@inproceedings{raistrick2023infinite,
  title={Infinite photorealistic worlds using procedural generation},
  author={Raistrick, Alexander and Lipson, Lahav and Ma, Zeyu and Mei, Lingjie and Wang, Mingzhe and Zuo, Yiming and Kayan, Karhan and Wen, Hongyu and Han, Beining and Wang, Yihan and others},
  booktitle={Proceedings of the IEEE/CVF conference on computer vision and pattern recognition},
  pages={12630--12641},
  year={2023}
}

@misc{rombach2021highresolution,
      title={High-Resolution Image Synthesis with Latent Diffusion Models}, 
      author={Robin Rombach and Andreas Blattmann and Dominik Lorenz and Patrick Esser and Björn Ommer},
      year={2021},
      eprint={2112.10752},
      archivePrefix={arXiv},
      primaryClass={cs.CV}
}

@misc{wang2026seeclearreliabletransparentobject,
      title={SeeClear: Reliable Transparent Object Depth Estimation via Generative Opacification}, 
      author={Xiaoying Wang and Yumeng He and Jingkai Shi and Jiayin Lu and Yin Yang and Ying Jiang and Chenfanfu Jiang},
      year={2026},
      eprint={2603.19547},
      archivePrefix={arXiv},
      primaryClass={cs.CV},
      url={https://arxiv.org/abs/2603.19547}, 
}

@inproceedings{liu2026foundationgeo,
  title={FoundationGeo: Learning Spatial Pixel-Wise Fields for Monocular Metric Geometry},
  author={Liu, Muxin and Lyu, Xiaoyang and Ren, Tianhe and Dai, Peng and Wu, Xiaoshan and Zhang, Zhiyue and Zhang, Jiaqi and Lin, Jiehong and Shi, Shaoshuai and Qi, Xiaojuan},
  booktitle={European Conference on Computer Vision},
  pages={353--371},
  year={2026},
  organization={Springer}
}

@inproceedings{lyu2026streamingdepth,
  title={Stabilizing streaming video geometry via dynamic feature normalization},
  author={Lyu, Xiaoyang and Liu, Muxin and Wu, Xiaoshan and Wang, Ruicheng and Huang, Yi-Hua and Sun, Yang-Tian and Shi, Shaoshuai and Qi, Xiaojuan},
  booktitle={Proceedings of the IEEE/CVF Conference on Computer Vision and Pattern Recognition},
  pages={7577--7587},
  year={2026}
}

@inproceedings{macenski2020marathon2,
  title     = {The Marathon 2: A Navigation System},
  author    = {Macenski, Steve and Martín, Francisco and White, Ruffin and Ginés Clavero, Jonatan},
  year      = {2020},
  booktitle = {2020 IEEE/RSJ International Conference on Intelligent Robots and Systems (IROS)},
  url       = {https://github.com/ros-planning/navigation2}
}

@inproceedings{lyu2021hr,
  title={Hr-depth: High resolution self-supervised monocular depth estimation},
  author={Lyu, Xiaoyang and Liu, Liang and Wang, Mengmeng and Kong, Xin and Liu, Lina and Liu, Yong and Chen, Xinxin and Yuan, Yi},
  booktitle={Proceedings of the AAAI conference on artificial intelligence},
  volume={35},
  number={3},
  pages={2294--2301},
  year={2021}
}

@misc{gao2025more3dvisualgeometry,
      title={MoRE: 3D Visual Geometry Reconstruction Meets Mixture-of-Experts}, 
      author={Jingnan Gao and Zhe Wang and Xianze Fang and Xingyu Ren and Zhuo Chen and Shengqi Liu and Yuhao Cheng and Jiangjing Lyu and Xiaokang Yang and Yichao Yan},
      year={2025},
      eprint={2510.27234},
      archivePrefix={arXiv},
      primaryClass={cs.CV},
      url={https://arxiv.org/abs/2510.27234}, 
}

@article{lu2025uniugp,
  title={Uniugp: Unifying understanding, generation, and planing for end-to-end autonomous driving},
  author={Lu, Hao and Liu, Ziyang and Jiang, Guangfeng and Luo, Yuanfei and Chen, Sheng and Zhang, Yangang and Chen, Ying-Cong},
  journal={arXiv preprint arXiv:2512.09864},
  year={2025}
}

@inproceedings{pei2026advancing,
  title={Advancing multi-agent traffic simulation via r1-style reinforcement fine-tuning},
  author={Pei, Muleilan and Shi, Shaoshuai and Shen, Shaojie},
  booktitle={International Conference on Learning Representations},
  volume={2026},
  pages={130791--130807},
  year={2026}
}

@inproceedings{jiang2026wpt,
  title={Wpt: World-to-policy transfer via online world model distillation},
  author={Jiang, Guangfeng and Luo, Yueru and Liu, Jun and Huang, Yi and Zhu, Yiyao and Qu, Zhan and Chen, Dave Zhenyu and Liu, Bingbing and Yan, Xu},
  booktitle={Proceedings of the IEEE/CVF Conference on Computer Vision and Pattern Recognition},
  pages={17842--17852},
  year={2026}
}

@inproceedings{tan2025xtrack,
  title={Xtrack: Multimodal training boosts rgb-x video object trackers},
  author={Tan, Yuedong and Wu, Zongwei and Fu, Yuqian and Zhou, Zhuyun and Sun, Guolei and Zamfir, Eduard and Ma, Chao and Paudel, Danda and Van Gool, Luc and Timofte, Radu},
  booktitle={2025 IEEE/CVF International Conference on Computer Vision (ICCV)},
  pages={5734--5744},
  year={2025},
  organization={IEEE}
}

@article{liu2026monocular,
  title={Monocular Depth Estimation from a Single Image: Progress and Opportunities},
  author={Liu, Muxin and Lyu, Xiaoyang and Sun, Yang-Tian and Huang, Yi-Hua and Yang, Ziyi and Dai, Peng and Qi, Xiaojuan},
  journal={arXiv preprint arXiv:2609.01172},
  year={2026}
}

@inproceedings{lin2024sam,
  title={Sam-6d: Segment anything model meets zero-shot 6d object pose estimation},
  author={Lin, Jiehong and Liu, Lihua and Lu, Dekun and Jia, Kui},
  booktitle={2024 IEEE/CVF Conference on Computer Vision and Pattern Recognition (CVPR)},
  pages={27906--27916},
  year={2024},
  organization={IEEE}
}
